%% file: main_cvpr.tex
\documentclass[10pt,letterpaper]{article}

\usepackage[pagenumbers]{cvpr} %

\usepackage{times}
\usepackage{mathptmx}
\usepackage{graphicx}
\usepackage{amsmath,amssymb}
\usepackage{booktabs}
\usepackage{makecell}
\usepackage{multirow}
\usepackage{xcolor}
\usepackage{colortbl}
\usepackage{pifont}
\usepackage{newunicodechar}
\usepackage[numbers]{natbib}  %
\usepackage{microtype}        %
\usepackage[font=small,labelfont=bf]{caption}

\newunicodechar{✓}{\textcolor{green!70!black}{\ding{51}}}
\newunicodechar{✗}{\textcolor{red}{\ding{55}}}
\newunicodechar{◯}{\textcolor{orange!90!black}{\textbf{O}}}

\definecolor{cvprblue}{rgb}{0.21,0.49,0.74}
\usepackage[pagebackref,breaklinks,colorlinks,allcolors=cvprblue]{hyperref}

\def\confName{CVPR}
\def\confYear{2026}

\title{Hand2World: Autoregressive Egocentric Interaction Generation via Free-Space Hand Gestures\vspace{-0.5em}}

\author{Yuxi Wang$^{1}$ \quad Wenqi Ouyang$^{1}$ \quad Tianyi Wei$^{1}$ \quad Yi Dong$^{1}$ \quad Zhiqi Shen$^{1*}$ \quad Xingang Pan$^{1*}$\\[0.20em]
$^{1}$College of Computing and Data Science, Nanyang Technological University\\
\scalebox{0.88}{\texttt{\{yuxiwang, wenqi.ouyang, tianyi.wei, yi.dong, zqshen, xingang.pan\}@ntu.edu.sg}}\\
}

\begin{document}
\flushbottom
\frenchspacing
\emergencystretch=1em

\twocolumn[{%
    \renewcommand\twocolumn[1][]{##1}%
    \maketitle
    \vspace{-3.0em}
    \begin{center}
        \captionsetup{type=figure}
        \includegraphics[width=\textwidth]{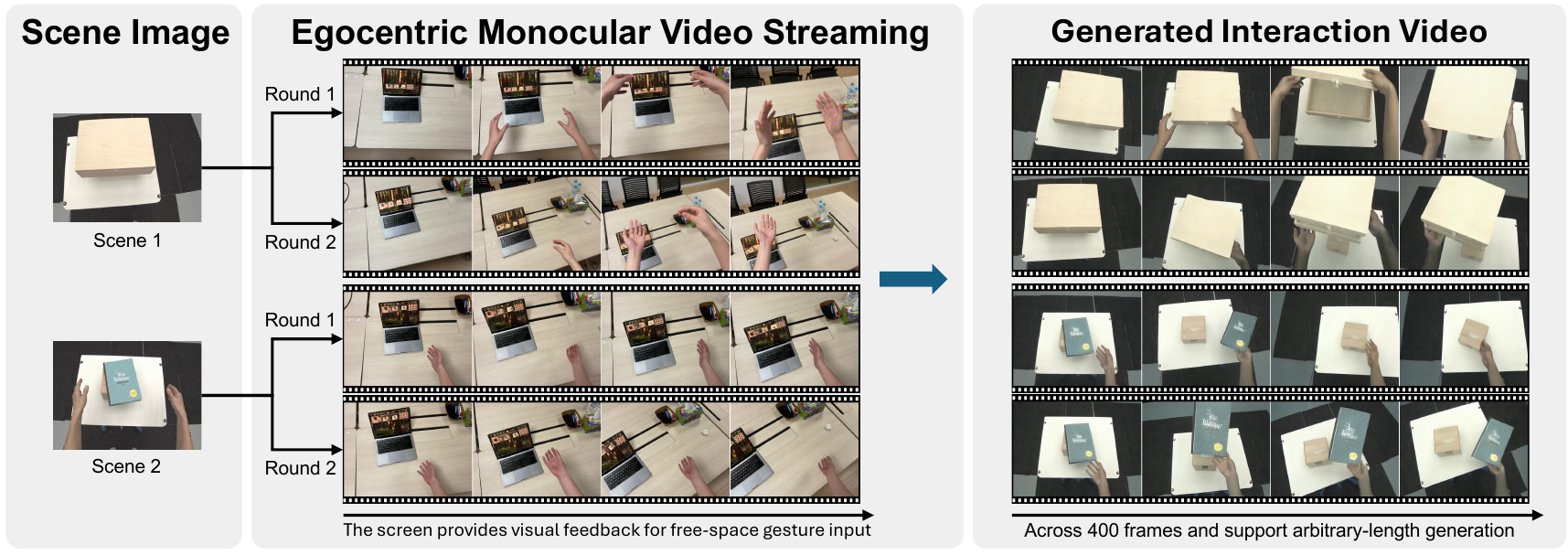}
        \captionof{figure}{Given a scene image (left) and free-space hand gestures from an egocentric monocular stream (middle), Hand2World synthesizes interaction videos (right) in which hands enter the depicted scene and manipulate objects while following the input viewpoint changes. Each scene is driven by two consecutive gesture rounds and generated autoregressively over 400+ frames, demonstrating arbitrary-length rollout.}
        \label{fig:teaser}
    \end{center}%
}]
\renewcommand{\thefootnote}{*}
\footnotetext{Corresponding authors.}

\begin{abstract}
\renewcommand{\thefootnote}{\fnsymbol{footnote}}
\setcounter{footnote}{0}
Egocentric interactive world models are essential for augmented reality and embodied AI, where visual generation must respond to user input with low latency, geometric consistency, and long-term stability. We study egocentric interaction generation from a single scene image under free-space hand gestures, aiming to synthesize photorealistic videos in which hands enter the scene, interact with objects, and induce plausible world dynamics under head motion. This setting introduces fundamental challenges, including distribution shift between free-space gestures and contact-heavy training data, ambiguity between hand motion and camera motion in monocular views, and the need for arbitrary-length video generation. We present Hand2World, a unified autoregressive framework that addresses these challenges through occlusion-invariant hand conditioning based on projected 3D hand meshes, allowing visibility and occlusion to be inferred from scene context rather than encoded in the control signal. To stabilize egocentric viewpoint changes, we inject explicit camera geometry via per-pixel Pl\"ucker-ray embeddings, disentangling camera motion from hand motion and preventing background drift. We further develop a fully automated monocular annotation pipeline and distill a bidirectional diffusion model into a causal generator, enabling arbitrary-length synthesis. Experiments on three egocentric interaction benchmarks show substantial improvements in perceptual quality and 3D consistency while supporting camera control and long-horizon interactive generation. Project page: \url{https://hand2world.github.io}.

\end{abstract}

\input{sec/1_intro}
\input{sec/2_related}
\input{sec/3_method}
\input{sec/4_experiments}
\input{sec/5_conclusion}

{
    \small
    \bibliographystyle{ieeenat_fullname}
    \bibliography{sample-base}
}

\input{sec/X_supplementary}

\end{document}

%% file: sec/1_intro.tex
\section{Introduction}

Interactive world models~\cite{ha2018world, bruce2024genie} offer a promising paradigm for simulating environment dynamics in augmented reality, teleoperation, and embodied AI. Among various control interfaces, hands provide one of the most natural signals: modern head-mounted devices can reliably capture \emph{free-space hand gestures}---contact-free motions performed in mid-air---enabling intuitive human-in-the-loop interaction in both virtual and real-world scenarios.

In this paper, we study \emph{egocentric interaction generation from a single scene image under free-space hand control}. Given a reference scene image and an egocentric gesture sequence, the goal is to synthesize a photorealistic video in which hands enter the depicted scene and manipulate objects as if the gestures were physically performed inside that environment. This requires a model that jointly addresses three fundamental challenges:
(i)~\textbf{Interaction plausibility}---hands must form realistic contacts with objects and induce physically plausible object responses;
(ii)~\textbf{Occlusion correctness}---hands should be correctly visible or occluded relative to scene objects, maintaining accurate depth ordering;
(iii)~\textbf{World consistency under head motion}---the natural ego-motion of the wearer should be reflected coherently in the background throughout the sequence.
Beyond these, practical deployment further requires \textbf{arbitrary-length, streaming-friendly generation}, which we refer to as \textbf{autoregressive} (\textbf{AR}) synthesis.

A scalable direction is to learn egocentric interactions from abundant monocular videos using deep generative models~\cite{wan2025}. Recent monocular approaches typically condition generation on 2D hand masks~\cite{coshand, interdyn, li2026mask2iv}, and have demonstrated strong performance for \emph{replaying} contact-heavy interaction videos. However, they face fundamental limitations under our target setting:
(i)~\textbf{Brittle free-space control.} Mask-based conditioning suffers from a severe \emph{distribution shift}: training masks are often partially visible due to occlusions, whereas free-space gestures are fully visible by construction. This conflation of \emph{hand geometry} with \emph{visibility state} causes the learned distribution to break down under complete masks, producing phantom occluders and incorrect depth ordering (Fig.~\ref{fig:mask_shift}).
(ii)~\textbf{Unstable scene geometry.} Egocentric videos are dominated by head-induced camera motion. Without explicit camera conditioning, models must infer viewpoint changes from appearance alone, leading to drifting backgrounds. Multi-view systems~\cite{tu2025, kim2025dwm} alleviate this but rely on synchronized rigs or scanned assets, limiting scalability.
(iii)~\textbf{Lack of online interaction pipelines.}
Closed-loop deployment demands a causal, end-to-end pipeline that ingests continuous monocular signals and produces frames in a streaming fashion. Such a pipeline, where a user drives scene interaction through live hand gestures and ego-motion, has not been explored.

These limitations suggest three design principles: \textbf{(P1)}~Occlusion should be \emph{inferred} from 3D geometry rather than prescribed by masks; \textbf{(P2)}~Camera motion must be \emph{explicit} to prevent background drift; and \textbf{(P3)}~Synthesis must be \emph{causal and streaming} for online deployment.

Building on these insights, we propose \textbf{Hand2World}, a unified framework for photorealistic egocentric interaction generation with free-space hand control, explicit camera control, and AR synthesis.
First, to eliminate mask distribution shift, we condition the generator on an \emph{occlusion-invariant hand representation}: we reconstruct 3D hand meshes (MANO~\cite{MANO2017}) and project their complete geometry to the image plane as a composite silhouette-and-wireframe signal. Unlike 2D masks, this representation remains \emph{format-consistent} regardless of occlusion state, allowing the model to learn occlusion reasoning from visual context rather than from visibility hard-coded into the condition.
Second, we inject \emph{explicit camera motion} using per-pixel Pl\"ucker-ray embeddings via a lightweight adapter, disentangling viewpoint changes from hand motion and stabilizing scene geometry under head movement.
Third, we develop a fully automated monocular annotation pipeline that extracts temporally consistent hand meshes and approximate camera trajectories from raw videos, avoiding manual labels and multi-view rigs.
Finally, we distill the bidirectional diffusion teacher into a causal AR generator~\cite{yin2025causvid, huang2025selfforcing}, enabling streaming-friendly, arbitrary-length synthesis.

Extensive experiments on three egocentric datasets~\cite{fan2023arctic, Banerjee_2025_CVPR, Liu_2022_CVPR} demonstrate that Hand2World substantially improves perceptual quality and 3D/viewpoint consistency, achieving a \textbf{76\%} reduction in FVD and a \textbf{42\%} reduction in camera trajectory error over state-of-the-art baselines (Tables~\ref{tab:comparison} and~\ref{tab:main_results}). Our main contributions are:
\begin{itemize}
  \item \textbf{The first monocular framework} that synthesizes photorealistic egocentric hand--object interaction videos from a single scene image, driven by unconstrained free-space gestures, with explicit camera control and arbitrary-length AR synthesis.
  \item \textbf{Occlusion-invariant hand conditioning} via projected 3D hand mesh controls (silhouette-and-wireframe) that encode complete hand geometry independent of visibility, eliminating mask distribution shift.
  \item \textbf{Explicit camera modeling} through Pl\"ucker-ray embeddings injected into the video diffusion model via a lightweight adapter, disentangling camera motion from hand motion and preventing background drift.
  \item \textbf{Scalable training and deployment:} a fully automated monocular annotation pipeline and autoregressive distillation, validated across three datasets.
\end{itemize}

\begin{table}[t]
  \centering
  \small
  \setlength{\tabcolsep}{2.5pt}
  \caption{Systematic comparison with state-of-the-art methods. \textbf{Free-G}: robust free-space gesture control. \textbf{Cam}: explicit camera control. \textbf{AR}: autoregressive generation. \textbf{Gen-Dyn}: generative scene dynamics. \textbf{Scale}: monocular training scalability. \textbf{Mono}: monocular inference. ◯ denotes limited support.}
  \label{tab:comparison}
  \arrayrulecolor{black}
  \resizebox{\columnwidth}{!}{%
  \begin{tabular}{lcccccc}
    \toprule
    \textbf{Method} & \textbf{Free-G} & \textbf{Cam} & \textbf{AR} & \textbf{Gen-Dyn} & \textbf{Scale} & \textbf{Mono} \\
    \midrule
    CosHand~\cite{coshand} & ◯ & ✗ & ✗ & ◯ & ✓ & ✓ \\
    InterDyn~\cite{interdyn} & ◯ & ✗ & ✗ & ✓ & ✓ & ✓ \\
    Mask2IV~\cite{li2026mask2iv} & ◯ & ✗ & ✗ & ◯ & ✓ & ✓ \\
    Re-HOLD~\cite{fan2025ReHOLD} & ✗ & ✗ & ✗ & ◯ & ✓ & ✓ \\
    SpriteHand~\cite{li2025spritehand} & ✗ & ◯ & ✓ & ✗ & ✓ & ✓ \\
    PlayerOne~\cite{tu2025} & ✓ & ◯ & ✓ & ✓ & ✗ & ✗ \\
    DWM~\cite{kim2025dwm} & ✓ & ◯ & ✗ & ✓ & ✗ & ✗ \\
    \midrule
    \rowcolor{yellow!30}\textbf{Hand2World (Ours)} & \textbf{✓} & \textbf{✓} & \textbf{✓} & \textbf{✓} & \textbf{✓} & \textbf{✓} \\
    \bottomrule
  \end{tabular}%
  }
  \vspace{-2.0em}
\end{table}

%% file: sec/2_related.tex
\section{Related Work}

\subsection{Generative World Models}
World models~\cite{ha2018world, lecun2022path} learn to predict environmental dynamics for planning and simulation. Recent work spans driving~\cite{hu2023gaia}, robotics~\cite{yang2023learning}, and interactive gaming; Genie~\cite{bruce2024genie}, GameNGen~\cite{valevski2024diffusion}, and Oasis~\cite{decart2024oasis} generate playable environments in real-time. Foundation models like Cosmos~\cite{agarwal2025cosmos} scale to millions of hours of video for physical AI. However, existing approaches either lack fine-grained hand control or require specialized hardware. Our work targets egocentric hand--object manipulation, learning from monocular video.
\subsection{Hand-Object Interaction Synthesis}
Synthesizing realistic hand interactions is a central challenge in computer vision. Table~\ref{tab:comparison} provides a systematic comparison across six dimensions. We discuss each in turn, following the order of challenges presented in the Introduction.

\noindent\textbf{Free-Space Gesture Control.}
Mask-based monocular methods~\cite{interdyn, li2026mask2iv, coshand} suffer from \emph{mask distribution shift}: training on contact-heavy data yields partial masks (hands frequently occluded), while free-space inference presents complete masks---a distribution mismatch causing phantom occluder artifacts. Multi-view methods~\cite{tu2025, kim2025dwm} achieve robust free-space control via 3D supervision, but require expensive capture rigs. Reenactment approaches~\cite{fan2025ReHOLD} require driving videos showing the target interaction, precluding free-space input. SpriteHand~\cite{li2025spritehand} renders hands over static backgrounds, avoiding occlusion reasoning entirely but forfeiting realistic scene integration.

\noindent\textbf{Camera Control.}
Monocular mask-based methods (CosHand, InterDyn, Mask2IV, Re-HOLD) lack explicit camera modeling, conflating hand motion with viewpoint changes and producing ``floating'' backgrounds that drift independently of intended camera motion. PlayerOne~\cite{tu2025} estimates egocentric camera from exocentric (third-person) views, introducing systematic bias due to the fundamental viewpoint discrepancy between ego and exo perspectives. DWM~\cite{kim2025dwm} and SpriteHand~\cite{li2025spritehand} support limited camera control but without explicit geometric grounding via Pl\"ucker embeddings.

\noindent\textbf{Arbitrary-Length Generation.}
Most video diffusion methods generate fixed-length clips due to bidirectional attention. Only PlayerOne~\cite{tu2025} and SpriteHand~\cite{li2025spritehand} support autoregressive generation for arbitrary-length synthesis; other methods are limited to short fixed-length clips or require sliding-window inference prone to boundary artifacts.

\noindent\textbf{Generative Scene Dynamics.}
CosHand~\cite{coshand} generates single images rather than videos, precluding temporal dynamics. Mask2IV~\cite{li2026mask2iv} requires pre-specified object trajectory masks as input rather than generating object motion, limiting generative capability. Re-HOLD~\cite{fan2025ReHOLD} transfers existing motions from driving videos rather than synthesizing novel dynamics. SpriteHand~\cite{li2025spritehand} renders hands over static backgrounds without object state changes. In contrast, InterDyn~\cite{interdyn}, PlayerOne~\cite{tu2025}, and DWM~\cite{kim2025dwm} generate full interaction dynamics including object motion.

\noindent\textbf{Scalability and Monocular Inference.}
Multi-view methods (PlayerOne, DWM) require synchronized ego-exocentric camera rigs or pre-scanned 3D assets for both training and inference, severely limiting scalability to diverse scenes and practical deployment. By contrast, monocular methods (CosHand, InterDyn, Mask2IV, Re-HOLD, SpriteHand) train on abundant single-view video and require only monocular input at inference, enabling in-the-wild deployment without specialized hardware.
\vspace{0.3em}
\noindent Our method is the first to achieve all six capabilities: robust free-space control via occlusion-invariant 3D hand conditioning, explicit camera control, autoregressive generation for arbitrary-length video, generative scene dynamics, and full monocular scalability for both training and deployment.
\subsection{Controllable Video Diffusion}
 ControlNet~\cite{zhang2023controlnet} and its video extensions~\cite{wang2023videocomposer} standardize spatial conditioning for diffusion models. Egocentric video additionally requires camera control to distinguish head motion from hand and object motion. Implicit camera learning~\cite{wang2024motionctrl} often produces ``floating background'' artifacts; we instead adopt explicit Pl\"ucker-ray embeddings~\cite{he2024cameractrl, plucker1865geometry} ensuring 3D-consistent viewpoint changes.
\subsection{Hand Modeling and Autoregressive Generation}
Monocular hand mesh estimation~\cite{HaMeR, yu2025dynhamr} enables extracting 3D hand geometry from unlabeled video. For arbitrary-length video synthesis, autoregressive distillation~\cite{yin2025causvid} with self-forcing~\cite{huang2025selfforcing} compresses bidirectional diffusion into causal generators, enabling arbitrary-length generation.

%% file: sec/3_method.tex
\section{Method}
\label{sec:method}

\subsection{Overview}
\label{sec:overview}

\begin{figure*}[t]
  \centering
  \includegraphics[width=\textwidth]{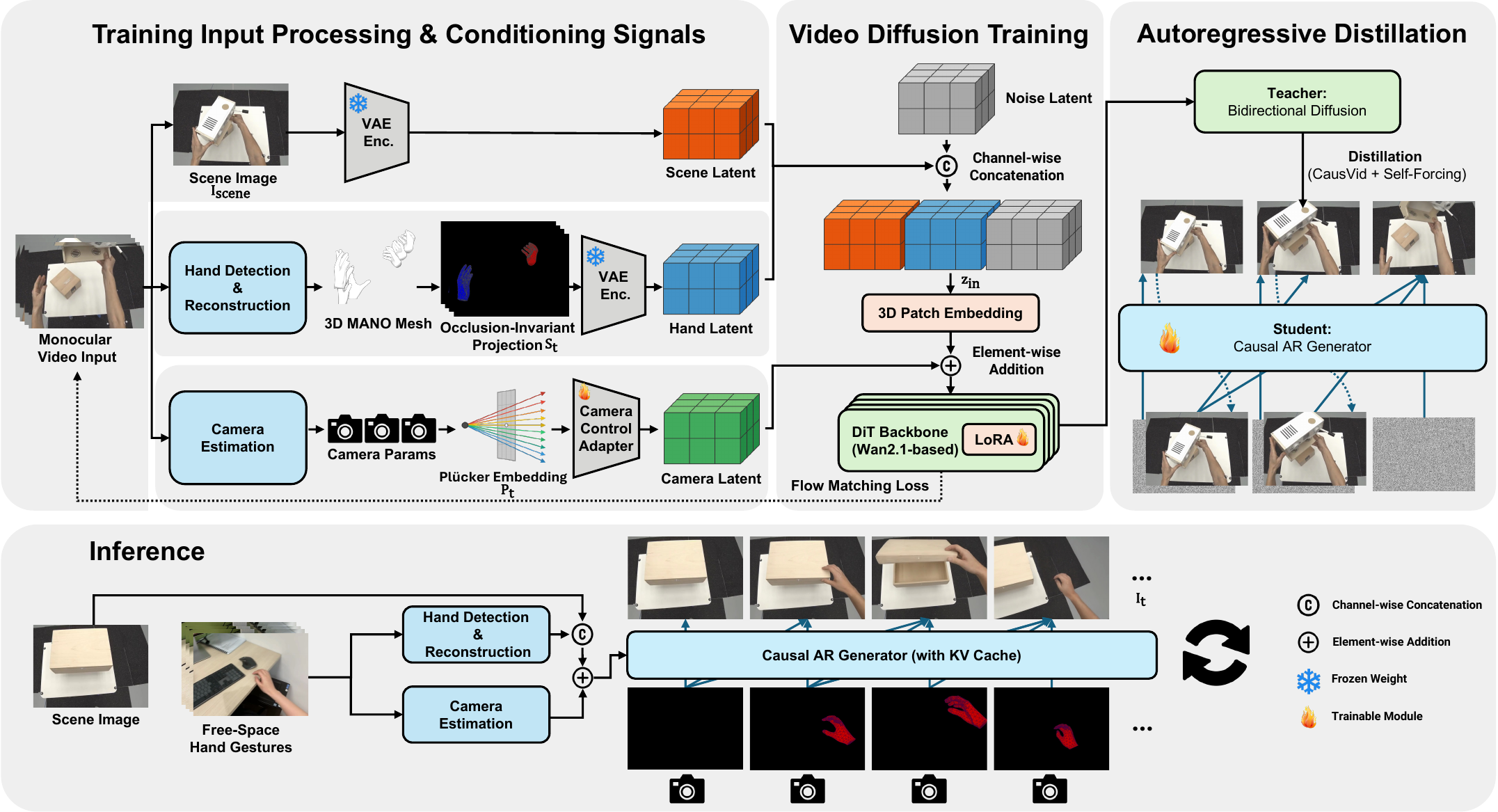}
  \caption{Hand2World encodes gestures as projected 3D hand meshes (silhouette-and-wireframe) to decouple geometry from visibility, and models camera motion via per-pixel Pl\"ucker-ray embeddings (left). During training (center), scene and hand controls are channel-concatenated into the diffusion transformer backbone (Wan2.1-1.3B-Control) while camera embeddings are injected additively through a lightweight adapter. The bidirectional teacher is distilled into a causal AR generator via CausVid~\cite{yin2025causvid} and self-forcing~\cite{huang2025selfforcing} (right). At inference (bottom), monocular estimators (HaMeR~\cite{HaMeR}, Depth Anything V3~\cite{depthanything3}) and KV-cached block-wise generation enable arbitrary-length egocentric interaction from a single scene image and free-space gestures.}
  \label{fig:pipeline}
  \vspace{-1.0em}
\end{figure*}

\begin{figure}[!htbp]
  \centering
  \includegraphics[width=\linewidth]{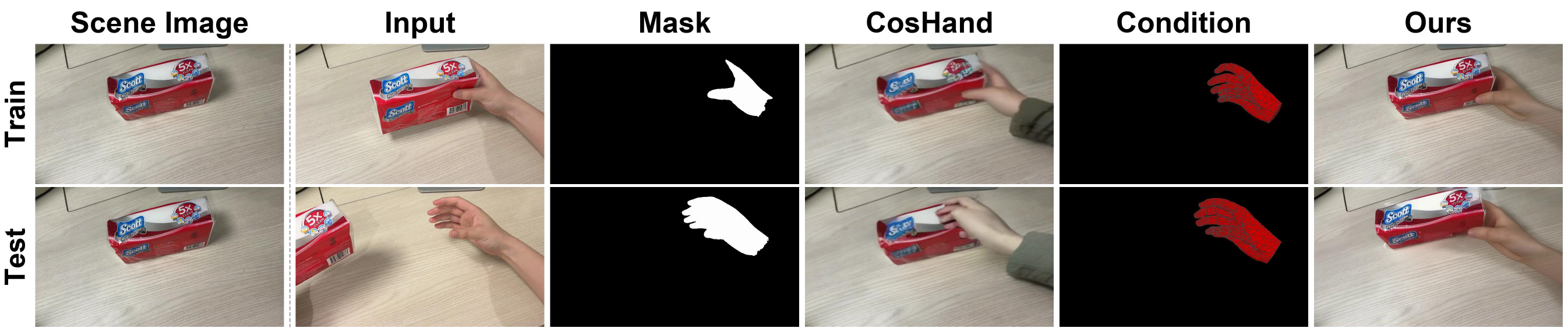}
  \caption{Training videos contain partially visible hands due to object occlusions (top), producing partial masks at training time. At inference, free-space gestures yield fully visible hands and complete masks (bottom), creating a train--test distribution gap. Mask-conditioned methods such as CosHand degrade under this shift, whereas our projected 3D hand mesh maintains a format-consistent control signal regardless of occlusion state.}
  \label{fig:mask_shift}
  \vspace{-1.0em}
\end{figure}

Hand2World generates an egocentric interaction video from a single reference scene image and a free-space hand-gesture sequence (Fig.~\ref{fig:pipeline}). Two observations specific to our setting motivate its core designs. First, mask-based hand conditioning conflates \emph{hand geometry} with \emph{visibility}: contact-heavy training videos contain partially visible hands, whereas free-space inference inputs are fully visible, creating a train--test mismatch that yields incorrect depth ordering (Fig.~\ref{fig:mask_shift}). Second, egocentric head motion produces substantial viewpoint changes that generators tend to entangle with hand motion, causing background drift. Hand2World addresses these issues with three key designs: \textbf{(1)~Occlusion-Invariant Hand Conditioning} (Sec.~\ref{sec:hand}), which represents gestures as projected 3D hand meshes to provide a visibility-independent control signal; \textbf{(2)~Explicit Camera Control} (Sec.~\ref{sec:camera}), which injects dense Pl\"ucker-ray embeddings to decouple viewpoint change from hand motion; and \textbf{(3)~Autoregressive Generator Distillation} (Sec.~\ref{sec:ar}), which distills the bidirectional diffusion teacher into a causal generator for arbitrary-length deployment. We build on Wan2.1-1.3B-Control~\cite{wan2025}, a flow-matching video diffusion transformer operating in a VAE-compressed latent space ($8{\times}$ spatial and $4{\times}$ temporal downsampling, $C{=}16$ channels).

\subsection{Problem Formulation}
\label{sec:formulation}

Let $I_{\text{scene}}$ denote a reference scene image, $\{H_t\}_{t=1}^{N}$ a hand gesture sequence encoding hand configurations at each frame, and $\{C_t\}_{t=1}^{N}$ a camera trajectory encoding viewpoint changes. Our goal is to learn a generator $G_\theta$ that synthesizes an egocentric video
\begin{equation}
  \{I_t\}_{t=1}^{N} \;=\; G_\theta\!\left(I_{\text{scene}},\; \{H_t\}_{t=1}^{N},\; \{C_t\}_{t=1}^{N}\right),
\end{equation}
such that (i)~hands enter the scene and induce plausible object dynamics, (ii)~depth ordering and occlusions are consistent with scene geometry, and (iii)~the background evolves consistently with the intended camera motion.

\subsection{Occlusion-Invariant Hand Conditioning}
\label{sec:hand}

Most prior work conditions on 2D visible-pixel masks extracted from training frames. However, when a hand grasps an object, parts of the hand become occluded and the corresponding mask shrinks. At inference, the same gesture performed in free space produces a \emph{complete} mask, creating a distribution gap that can confuse the generator into hallucinating occluders or misjudging depth ordering. To sidestep this issue, we design a hand control signal that specifies intended \emph{3D geometry} while leaving \emph{visibility} to be inferred from the scene context.

We parameterize each gesture as $H_t^h=(\beta^h,\theta_t^h,\mathbf{t}_t^h)$ for hand $h\in\{L,R\}$ (MANO~\cite{MANO2017} shape, pose, translation) and transform it into an occlusion-invariant control signal $S_t$ as the actual generator input.

\noindent\textbf{Hand mesh construction.}
For each hand at time~$t$, we compute 3D vertices via the parametric hand model $M$:
\begin{equation}
  V_t^h \;=\; M(\beta^h,\theta_t^h) + \mathbf{t}_t^h \;\in\; \mathbb{R}^{778\times 3}.
\end{equation}

\noindent\textbf{Projection and rendering.}
We project the vertices onto the image plane using camera intrinsics $K_t$ and rasterize them into a two-layer composite control frame:
\begin{equation}
  S_t \;=\; \mathrm{Render}\!\left(V_t^L,\, V_t^R;\, K_t\right).
\end{equation}
The first layer is a filled \emph{silhouette} that constrains spatial extent and coarse hand shape; the second is a \emph{wireframe} overlay that exposes articulation structure, namely finger configuration and joint topology. Left and right hands are color-coded to preserve identity under bimanual overlap. Compared to silhouettes alone, the wireframe provides additional articulation cues under self-occlusion (\eg, palm-facing poses), improving adherence to fine-grained gestures (Sec.~\ref{sec:ablation}).

Crucially, because this representation is derived from a 3D mesh rather than from observed pixels, it does not encode any visibility convention (\eg, ``only visible pixels'') and therefore remains \emph{format-consistent} regardless of whether training data contain contact-heavy occlusions or inference inputs are free-space gestures. Occlusion reasoning is thus delegated to the generator, which must infer depth ordering from the scene context.

\subsection{Explicit Camera Control}
\label{sec:camera}

Egocentric head motion produces global parallax and background shifts that are not identifiable from hand motion alone. Without explicit supervision, we observe that generators tend to ``explain away'' viewpoint change by warping the background in hand-correlated patterns, resulting in drift. We therefore condition the generator on per-pixel Pl\"ucker-ray embeddings derived from the camera trajectory $\{C_t\}$, providing a dense, spatially grounded viewpoint signal.

\noindent\textbf{Pl\"ucker-ray parameterization.}
Each camera parameter $C_t=(R_t,\mathbf{t}_t,K_t)$ specifies rotation, translation, and intrinsics. For each pixel $(u,v)$ at time~$t$, we compute:
\begin{equation}
  \mathbf{d}_{t}(u,v) = \frac{R_t^\top K_t^{-1}[u,v,1]^\top}{\|R_t^\top K_t^{-1}[u,v,1]^\top\|_2},\qquad
  \mathbf{o}_{t} = -R_t^\top \mathbf{t}_t,
\end{equation}
and form the 6D Pl\"ucker-ray coordinates as
\begin{equation}
  P_t(u,v) \;=\; \bigl(\mathbf{m}_t(u,v),\;\mathbf{d}_t(u,v)\bigr), \qquad \mathbf{m}_t(u,v)=\mathbf{d}_t(u,v)\times \mathbf{o}_t.
\end{equation}

We describe how these Pl\"ucker-ray maps are injected into the diffusion backbone in Sec.~\ref{sec:arch}.

\subsection{Monocular Auto-Annotation Pipeline}
\label{sec:annotation}

Training our model requires paired supervision of hand geometry and camera motion for each video frame, but such annotations are generally unavailable for in-the-wild egocentric footage. We therefore build an automatic annotation pipeline that recovers both modalities from monocular video and uses them as pseudo-labels for conditioning.

\noindent\textbf{Hand mesh extraction and temporal stabilization.}
We run a YOLO-based hand detector per frame and apply lightweight temporal heuristics: IoU-based duplicate removal, boundary suppression for boxes within 10\% of the image edge, and linear interpolation across short missing segments. MANO parameters are then estimated with HaMeR~\cite{HaMeR} and rendered into silhouette-and-wireframe controls (Sec.~\ref{sec:hand}). These steps improve temporal stability and reduce flicker in the control stream (Sec.~\ref{sec:ablation}).

\noindent\textbf{Camera trajectory estimation.}
We estimate per-frame camera parameters $(R_t,\mathbf{t}_t,K_t)$ from monocular video and normalize all trajectories relative to the first frame. Depth is used only to support pose recovery and is \emph{not} provided to the generator at any stage.

\subsection{Dual-Pathway Video Diffusion}
\label{sec:arch}

With the hand and camera conditioning pathways in place, we now describe how they are integrated into the video diffusion backbone. Following the principle of \emph{disentangled information injection}, which fuses different types of conditioning through structurally distinct pathways, we feed hand control and the scene image as dense latent channels, while camera geometry modulates the token stream through additive injection.

\noindent\textbf{Latent construction.}
Let $\mathbf{z}^{(\tau)}\in\mathbb{R}^{C\times T\times H'\times W'}$ denote the noisy video latent at diffusion timestep $\tau$, where $T=\lceil N/4\rceil$, $H'=H/8$, and $W'=W/8$. We encode the hand-control video $\{S_t\}_{t=1}^{N}$ into a latent of the same spatial and temporal resolution:
\begin{equation}
  \mathbf{z}_h \;=\; \mathrm{Enc}\bigl(\{S_t\}_{t=1}^{N}\bigr) \;\in\; \mathbb{R}^{C\times T\times H'\times W'}.
\end{equation}
The reference scene image is encoded only at the first latent time index, with the remaining indices zero-padded:
\begin{equation}
  \mathbf{z}_r \;=\; \bigl[\mathrm{Enc}(I_{\text{scene}}),\, \mathbf{0},\, \ldots,\, \mathbf{0}\bigr]
  \;\in\; \mathbb{R}^{C\times T\times H'\times W'}.
\end{equation}
These three latents are then concatenated channel-wise to form the input:
\begin{equation}
  \mathbf{z}_{\text{in}} \;=\; \bigl[\mathbf{z}^{(\tau)};\;\mathbf{z}_h;\;\mathbf{z}_r\bigr] \;\in\; \mathbb{R}^{3C\times T\times H'\times W'}.
\end{equation}
\noindent\textbf{Camera injection.}
Rather than concatenating camera features channel-wise, which would increase dimensionality and risk entangling camera cues with appearance, we inject them additively. A small adapter $a_{\text{cam}}$ temporally packs consecutive Pl\"ucker-ray maps $\{P_t\}$ (Sec.~\ref{sec:camera}) to match the VAE's $4{\times}$ temporal stride, downsamples them to the patch grid, and projects the 6D ray features to the token embedding dimension:
\begin{equation}
  \mathbf{h}_0 \;=\; \mathrm{Emb}_{\text{patch}}(\mathbf{z}_{\text{in}}) \;+\; a_{\text{cam}}(\{P_t\}).
\end{equation}
This additive design provides a dedicated pathway for viewpoint cues, allowing the model to attribute global parallax to the explicit camera signal rather than inferring it from hand motion, reducing background drift (Sec.~\ref{sec:ablation}). The transformer predicts flow-matching velocity fields for denoising.

\noindent\textbf{Training strategy.}
The model is trained with the rectified flow objective~\cite{liu2022flow}:
\begin{equation}
  \mathcal{L} \;=\; \mathbb{E}_{\tau,\,\mathbf{z}_0,\,\boldsymbol{\epsilon}}\!\left[\left\| v_\theta\!\left(\mathbf{z}_{\text{in}},\,\tau,\,\{P_t\}\right) - \left(\boldsymbol{\epsilon} - \mathbf{z}_0\right) \right\|^2\right],
\end{equation}
where $\mathbf{z}^{(\tau)}=(1{-}\tau)\,\mathbf{z}_0 + \tau\,\boldsymbol{\epsilon}$ linearly interpolates between the data latent~$\mathbf{z}_0$ and Gaussian noise~$\boldsymbol{\epsilon}$.
Directly fine-tuning all parameters with the new conditioning pathways risks destabilizing the pretrained generative prior. We therefore adopt a two-stage schedule: we first pretrain the camera adapter with the transformer backbone frozen, allowing the adapter to learn meaningful viewpoint representations; we then jointly fine-tune the adapter together with low-rank adaptation (LoRA) modules inserted into the transformer. This staged approach stabilizes camera conditioning while preserving the base model's image and motion priors.

\subsection{Autoregressive Generator Distillation}
\label{sec:ar}

Bidirectional diffusion models are typically constrained to synthesizing clips of fixed duration. Interactive and real-world applications, however, demand streaming-friendly, long-horizon generation. To bridge this gap, we distill our bidirectional diffusion teacher into a causal autoregressive (AR) generator, following the CausVid paradigm~\cite{yin2025causvid} combined with self-forcing strategies~\cite{huang2025selfforcing}.

\noindent\textbf{Distillation procedure.}
We first initialize the student via ODE pretraining on teacher-generated trajectories, then apply distribution matching distillation~\cite{yin2024improved} to align its output distribution with that of the bidirectional teacher. To mitigate exposure bias caused by train-time reliance on teacher-provided context, we adopt self-forcing~\cite{huang2025selfforcing}, replacing history frames with the student's predictions during training.

\noindent\textbf{Block-wise inference.}
At inference time, the AR generator produces frames sequentially in blocks, caching key/value states as context for subsequent blocks. Compared to sliding-window inference, this block-wise strategy maintains temporal coherence at block boundaries and naturally supports arbitrary-length rollouts, enabling our system to synthesize extended egocentric interaction videos from a single scene image and free-space gestures.

%% file: sec/4_experiments.tex
\section{Experiments}
\label{sec:experiments}

We evaluate Hand2World on egocentric interaction generation under free-space hand control. Our evaluation targets three capabilities aligned with the problem motivation:
(i) \textbf{perceptual realism and temporal coherence} of synthesized videos,
(ii) \textbf{controllability} with respect to fine-grained hand articulation and camera motion,
and (iii) \textbf{3D/viewpoint consistency} under ego-motion, including long-horizon AR rollouts.
We report results for both the bidirectional diffusion teacher (Hand2World) and its distilled causal variant (Hand2World-AR), where \textbf{AR denotes autoregressive}.

\subsection{Experimental Setup}
\label{sec:exp_setup}

\noindent\textbf{Implementation.}
We build on Wan2.1-1.3B-Control~\cite{wan2025} with a two-stage training protocol.
We first pretrain the camera adapter for 10{,}000 steps while freezing the transformer backbone to align coordinate systems.
We then jointly fine-tune the adapter together with rank-256 LoRA~\cite{lora} modules inserted into the attention and feed-forward layers for 100{,}000 steps.
Training uses AdamW with bf16 mixed precision on 8$\times$ NVIDIA A100-80GB GPUs with a global batch size of 8.
We sample 41-frame clips resized to 480p on the short side (aspect ratio preserved), and the AR generator trains on 384p for efficiency.

\noindent\textbf{AR distillation.}
We distill the bidirectional model into an AR generator following CausVid~\cite{yin2025causvid} and self-forcing~\cite{huang2025selfforcing}.
The generator is initialized with ODE pretraining and refined using distribution matching distillation.

\noindent\textbf{Datasets and baselines.}
Our primary evaluation is conducted on ARCTIC~\cite{fan2023arctic}; additional results on HOT3D~\cite{Banerjee_2025_CVPR} and HOI4D~\cite{Liu_2022_CVPR} are reported in the supplementary material.
We compare against CosHand~\cite{coshand}, InterDyn~\cite{interdyn}, Mask2IV~\cite{li2026mask2iv}, and Wan2.1-1.3B-Control~\cite{wan2025} without our proposed hand/camera conditioning.
All methods use identical test splits and consistent preprocessing.
For mask-based baselines, we provide SAM3-derived~\cite{carion2025sam3segmentconcepts} segmentation masks to ensure high-quality hand control signals, isolating the effect of our occlusion-invariant hand representation.

We do not include DWM~\cite{kim2025dwm}, Re-HOLD~\cite{fan2025ReHOLD}, or SpriteHand~\cite{li2025spritehand} due to the lack of public implementations at submission time.
PlayerOne~\cite{tu2025} requires synchronized ego--allocentric capture and is not directly comparable in our monocular setting; we include it only in qualitative comparisons when results are available.

\noindent\textbf{Evaluation metrics.}
We report Fr\'echet Video Distance (FVD)~\cite{styleganv} for distributional realism and DINO similarity~\cite{dino} for semantic alignment.
PSNR, SSIM~\cite{wang2004ssim}, and LPIPS~\cite{lpips} measure frame-level fidelity.
Flow-ERR measures temporal coherence using RAFT~\cite{raft}.
Depth-ERR and Cam-ERR evaluate 3D structure and viewpoint consistency by comparing depth and camera trajectories estimated with Depth Anything V3~\cite{depthanything3}.
Metric definitions are in Sec.~\ref{sec:supp_metrics}.

\subsection{Quantitative Results}
\label{sec:quant_results}

\begin{table*}[t]
  \centering
  \small
  \caption{\textbf{Quantitative comparisons on the ARCTIC test set.}
  We evaluate distributional realism (FVD, lower is better), semantic alignment (DINO, higher), frame-level fidelity (PSNR/SSIM higher; LPIPS lower), temporal coherence (Flow-ERR lower), and 3D/viewpoint consistency (Depth-ERR/Cam-ERR lower).
  Hand2World achieves the strongest overall results, and the distilled Hand2World-AR remains close while enabling causal long-horizon generation. Best results are \textbf{bold}.}
  \label{tab:main_results}
  \setlength{\tabcolsep}{6pt}
  \begin{tabular}{lcccccccc}
    \toprule
    \textbf{Method} & \textbf{FVD}$\downarrow$ & \textbf{DINO}$\uparrow$ & \textbf{PSNR}$\uparrow$ & \textbf{SSIM}$\uparrow$ & \textbf{LPIPS}$\downarrow$ & \textbf{Flow-ERR}$\downarrow$ & \textbf{Depth-ERR}$\downarrow$ & \textbf{Cam-ERR}$\downarrow$ \\
    \midrule
    CosHand~\cite{coshand} & 1400.68 & 0.79 & 12.41 & 0.63 & 0.55 & 78.87 & 22.51 & 0.14 \\
    InterDyn~\cite{interdyn} & 908.32 & 0.80 & 12.43 & 0.63 & 0.53 & 72.20 & 27.23 & 0.13 \\
    Mask2IV~\cite{li2026mask2iv} & 1083.62 & 0.78 & 11.47 & 0.62 & 0.55 & 70.45 & 31.34 & 0.12 \\
    Wan2.1-1.3B-Control~\cite{wan2025} & 1038.53 & 0.75 & 12.81 & 0.62 & 0.49 & 63.71 & 22.66 & 0.13 \\
    \midrule
    \rowcolor{yellow!30}\textbf{Hand2World-AR (Ours)} & 232.40 & \textbf{0.88} & 14.55 & \textbf{0.67} & 0.41 & 46.10 & 17.02 & \textbf{0.07} \\
    \rowcolor{yellow!30}\textbf{Hand2World (Ours)} & \textbf{218.76} & \textbf{0.88} & \textbf{14.93} & \textbf{0.67} & \textbf{0.39} & \textbf{44.43} & \textbf{16.14} & \textbf{0.07} \\
    \bottomrule
  \end{tabular}
  \vspace{-1.0em}
\end{table*}

Table~\ref{tab:main_results} summarizes results on ARCTIC.
Hand2World improves both perceptual quality and controllability-related metrics over prior work.
Compared to InterDyn, the strongest baseline in terms of FVD, Hand2World reduces FVD from 908.32 to 218.76 and improves DINO similarity from 0.80 to 0.88.
It also improves 3D/viewpoint consistency, achieving lower Depth-ERR and Cam-ERR than all baselines (Table~\ref{tab:main_results}).
The distilled model Hand2World-AR remains close to the teacher at the 81-frame horizon while enabling efficient long-horizon rollouts (Sec.~\ref{sec:ablation_ar}).

\subsection{Ablation Study}
\label{sec:ablation}

We ablate key components on ARCTIC to validate the contributions of explicit camera conditioning, wireframe-augmented hand rendering, and annotation-time temporal stabilization (Table~\ref{tab:ablation}). Full ablations on all three datasets are in the supplementary material (Table~\ref{tab:supp_ablation_full}).

\begin{table}[t]
  \centering
  \small
  \caption{\textbf{Ablation study on ARCTIC.} Best results are \textbf{bold}. Removing explicit camera conditioning yields the largest degradation, while wireframe augmentation and temporal stabilization provide complementary gains.}
  \label{tab:ablation}
  \setlength{\tabcolsep}{3pt}
  \begin{tabular}{lcccc}
    \toprule
    \textbf{Configuration} & \textbf{FVD}$\downarrow$ & \textbf{DINO}$\uparrow$ & \textbf{PSNR}$\uparrow$ & \textbf{Cam-ERR}$\downarrow$ \\
    \midrule
    w/o Camera Adapter & 815.14 & 0.84 & 12.66 & 0.13 \\
    w/o Wireframe & 223.29 & 0.87 & 14.78 & 0.08 \\
    w/o Interpolation & 231.09 & 0.87 & 14.75 & 0.08 \\
    w/o Edge Filter & 219.15 & 0.87 & 14.76 & 0.08 \\
    w/o Overlap Filter & 219.75 & 0.87 & 14.78 & 0.08 \\
    \midrule
    \textbf{Full Model} & \textbf{218.76} & \textbf{0.88} & \textbf{14.93} & \textbf{0.07} \\
    \bottomrule
  \end{tabular}
  \vspace{-1.0em}
\end{table}

\begin{figure}[t]
  \centering
  \includegraphics[width=\linewidth]{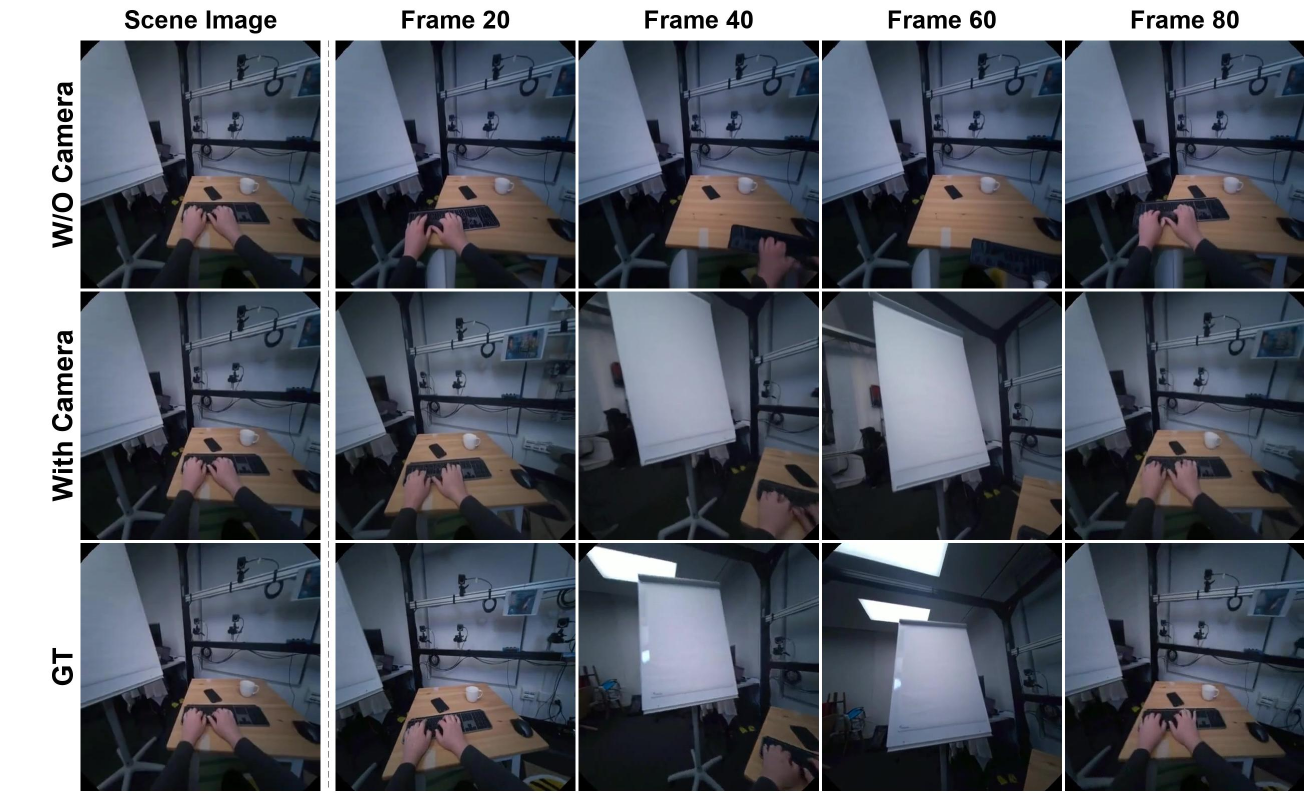}
  \caption{Without explicit camera conditioning (top), the model fails to follow the rightward-then-leftward head pan and produces background drift. With Pl\"ucker-ray injection (middle), the generated trajectory closely matches the ground truth (bottom).}
  \label{fig:ablation_camera}
  \vspace{-1.0em}
\end{figure}

Table~\ref{tab:ablation} decomposes the contributions of explicit camera conditioning, wireframe-augmented hand controls, and annotation-time temporal stabilization.
The camera adapter is the dominant factor: removing it increases FVD to 815.14 and raises Cam-ERR to 0.13.
This reflects the fact that global ego-motion is a primary source of variation in egocentric videos; without an identifiable camera signal, the model tends to entangle viewpoint change with scene dynamics, leading to background drift and unstable parallax (Fig.~\ref{fig:ablation_camera}).

\begin{figure}[t]
  \centering
  \includegraphics[width=\linewidth]{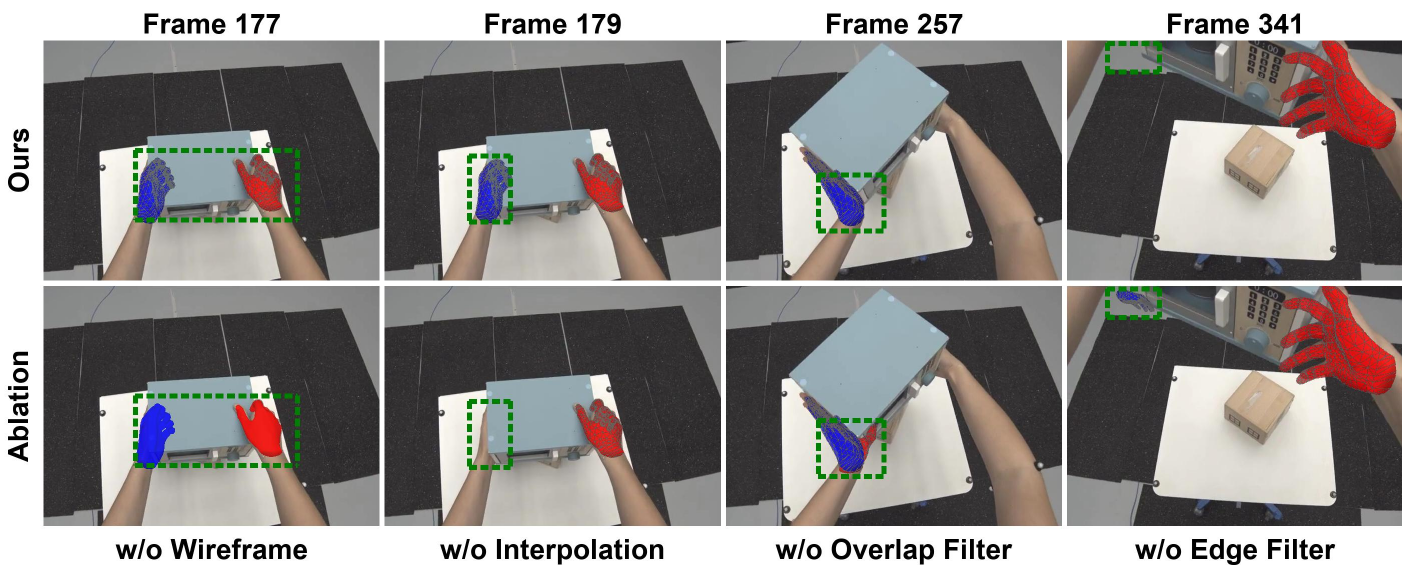}
  \caption{Each column compares the full annotation pipeline (top) against a single-component ablation (bottom) at a representative frame. Green boxes highlight affected regions.}
  \label{fig:detection_ablation}
  \vspace{-1.0em}
\end{figure}

On top of camera conditioning, wireframe augmentation and temporal stabilization provide complementary gains (Table~\ref{tab:ablation}).
The wireframe overlay supplies articulation cues that can be ambiguous from a silhouette under self-occlusion, especially when the palm faces the camera and fingers are largely occluded. This helps the generator maintain plausible finger configurations.
Temporal interpolation and filtering reduce brief missing or misaligned segments in per-frame detections, which otherwise manifest as flicker or discontinuous motion for input conditions (Fig.~\ref{fig:detection_ablation}).

\begin{figure*}[!t]
  \centering
  \includegraphics[width=\textwidth]{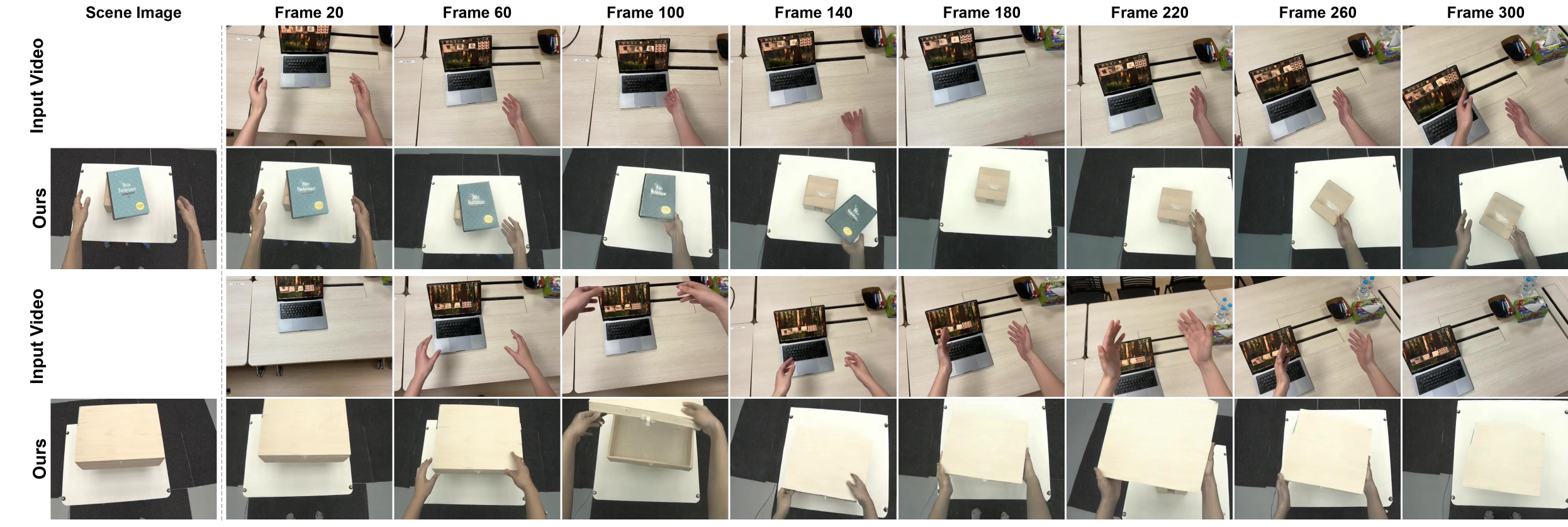}
  \caption{For each example, the input gesture video (odd rows) provides the driving hand motion, while the scene image (leftmost column) defines the target environment; Hand2World-AR generates the interaction sequences (even rows).
  Rows~1--2: the model synthesizes sequential multi-object manipulation---grasping and removing a thin book to reveal the box underneath, then pushing the box---while following the ego-motion present in the input stream and maintaining a stable background.
  Rows~3--4: starting from a closed box, the model opens the lid to reveal a plausible interior, then closes and repositions the box, preserving object identity and geometric consistency.}

  \label{fig:freespace_ours_long}
  \vspace{-1.0em}
\end{figure*}

\subsection{Autoregressive Generation}
\label{sec:ablation_ar}

We distill the bidirectional teacher into a causal AR generator to enable streaming, arbitrary-length synthesis.
Table~\ref{tab:ar_length} evaluates Hand2World-AR as the rollout length increases from 81 to 324 frames.
At 81 frames, the AR generator closely matches the teacher. Quality degrades gradually at longer horizons as autoregressive errors accumulate, yet the model maintains coherent camera motion and plausible interaction dynamics throughout.
Hand2World-AR achieves 8.9 FPS on a single A100 GPU at $544{\times}384$ resolution. Detailed runtime is in Sec.~\ref{sec:supp_runtime}.

\begin{table}[t]
  \centering
  \small
  \caption{\textbf{Generation length ablation on ARCTIC.} Hand2World-AR maintains quality and degrades gradually at longer rollouts.}
  \label{tab:ar_length}
  \setlength{\tabcolsep}{3pt}
  \begin{tabular}{lccccc}
    \toprule
    \textbf{Attention} & \textbf{Len} & \textbf{FVD}$\downarrow$ & \textbf{DINO}$\uparrow$ & \textbf{PSNR}$\uparrow$ & \textbf{Cam-ERR}$\downarrow$ \\
    \midrule
    Bidirectional & 81 & \textbf{218.76} & \textbf{0.88} & \textbf{14.93} & \textbf{0.07} \\
    Autoregressive & 81 & 232.40 & \textbf{0.88} & 14.55 & \textbf{0.07} \\
    Autoregressive & 162 & 264.20 & \textbf{0.88} & 13.53 & 0.12 \\
    Autoregressive & 324 & 331.71 & 0.86 & 12.45 & 0.17 \\
    \bottomrule
  \end{tabular}
  \vspace{-1.0em}
\end{table}

\subsection{Qualitative Results}
\label{sec:qual_results}

Figure~\ref{fig:freespace_ours_long} presents long-horizon rollouts driven solely by free-space (mid-air) hand gestures.
Given a reference scene image and a monocular egocentric gesture stream, Hand2World-AR synthesizes interaction videos that remain stable under substantial head-induced viewpoint changes, while maintaining consistent hand--object occlusions during contact.

In the first example (top), the user manipulates a \emph{thin} book, which provides limited graspable thickness and therefore requires precise finger articulation and robust adherence to the input hand controls to lift without visual artifacts.
Notably, when the book rests on the box, the box is heavily occluded and only a small corner is visible.
Despite this partial observation, the model performs plausible amodal completion of the box and keeps its shape and appearance stable as the book is grasped, lifted, and removed, rather than letting the underlying object flicker or drift.
After removing the book, the rollout naturally transitions to interacting with the box, suggesting that the learned control is not tied to a single target object but supports sequential interactions with multiple objects in the same scene.

In the second example (bottom), the gesture stream alternates between motions that indicate opening the lid and motions that indicate lifting/carrying the box.
The model follows these changes by opening the hinged lid and synthesizing a plausible interior, and then closing and picking up the box while preserving its identity and temporal continuity throughout the rollout.
Additional qualitative comparisons are provided in the supplementary material.

%% file: sec/5_conclusion.tex
\section{Conclusion}
\label{sec:conclusion}

We introduced Hand2World, a monocular framework for generating egocentric interaction videos from a scene image and free-space hand gestures. Hand2World decouples hand articulation from visibility by conditioning on occlusion-invariant projections of 3D hand meshes, and improves viewpoint consistency by injecting explicit camera control via Pl\"ucker-ray embeddings. Combined with monocular auto-annotation and autoregressive distillation, the framework enables controllable long-horizon rollouts and achieves strong performance on egocentric interaction benchmarks.

\noindent\textbf{Limitations.}\quad Free-space gestures lack physical contact constraints, so users may specify infeasible motions (e.g., penetrating solid objects), potentially leading to implausible interactions. Incorporating force-feedback devices could help enforce physical plausibility.

\noindent\textbf{Acknowledgements.}\quad This research is supported by RIE2025 Industry Alignment Fund – Industry Collaboration Projects (IAF-ICP) (Award I2301E0026), administered by A*STAR, and by Alibaba Group and NTU Singapore through the Alibaba-NTU Global e-Sustainability CorpLab (ANGEL).

%% file: sec/X_supplementary.tex
\appendix
\onecolumn

\setcounter{topnumber}{50}
\setcounter{bottomnumber}{50}
\setcounter{totalnumber}{50}
\renewcommand{\topfraction}{0.95}
\renewcommand{\bottomfraction}{0.95}
\renewcommand{\textfraction}{0.05}
\renewcommand{\floatpagefraction}{0.90}
\setlength{\textfloatsep}{10pt plus 2pt minus 2pt}
\setlength{\floatsep}{8pt plus 2pt minus 2pt}
\setlength{\intextsep}{8pt plus 2pt minus 2pt}

\newcommand{\SuppFigure}[4][0.86]{%
  \begin{center}
  \begin{minipage}{\textwidth}
    \centering
    \includegraphics[width=\textwidth,height=#1\textheight,keepaspectratio]{\detokenize{#2}}
    \captionof{figure}{#3}
    \label{#4}
  \end{minipage}
  \end{center}
  \vspace{0.5em}
}

\begin{center}
{\Large\bfseries Supplementary Material}\\[2pt]
\end{center}

\vspace{0.5em}

\section*{Overview}

This supplementary provides additional details and results that support the main paper.
Sec.~\ref{sec:supp_impl} describes the backbone, conditioning design, and training/inference settings.
Sec.~\ref{sec:supp_pipeline} details our monocular auto-annotation pipeline for hand meshes and camera trajectories.
Sec.~\ref{sec:supp_ar} expands the autoregressive distillation procedure.
Sec.~\ref{sec:supp_eval} defines datasets and metrics, and reports extended quantitative comparisons and ablations.
Sec.~\ref{sec:supp_runtime} reports runtime measurements of the full pipeline (Table~\ref{tab:supp_runtime}).
Sec.~\ref{sec:supp_qual} presents qualitative comparisons (Figures~\ref{fig:supp_recon_arctic_box_open}--\ref{fig:supp_recon_hot3d_down}).
Sec.~\ref{sec:supp_challenge_case} presents qualitative results on challenging long-horizon cases (Figure~\ref{fig:supp_long}).
Sec.~\ref{sec:supp_freespace} reports free-space gesture generation under camera motion (Figures~\ref{fig:supp_freespace_1} and~\ref{fig:supp_freespace_2}).

\section{Implementation Details}
\label{sec:supp_impl}

\subsection{Backbone and Latent Representation}

We build upon \textbf{Wan2.1-1.3B-Control}~\cite{wan2025}, a flow-matching video diffusion transformer operating in a VAE latent space.
The VAE applies $8\times$ spatial and $4\times$ temporal downsampling, producing latent channels $C{=}16$.
Unless otherwise specified, we use an empty text prompt for all experiments to isolate the effect of hand and camera conditioning.

\subsection{Dual-Pathway Input Conditioning}

\noindent\textbf{Latents.}
Let $\mathbf{z}^{(\tau)} \in \mathbb{R}^{C \times T \times H' \times W'}$ denote the noisy video latent at diffusion timestep $\tau$.
We encode the rendered hand-control video $\{S_t\}_{t=1}^{N}$ into
\begin{equation}
\mathbf{z}_h = \mathrm{Enc}(\{S_t\}_{t=1}^{N}) \in \mathbb{R}^{C \times T \times H' \times W'}.
\end{equation}
We encode the reference scene image only at the first latent index and pad the rest with zeros:
\begin{equation}
\mathbf{z}_r = \big[\mathrm{Enc}(I_{\text{scene}}),\, \mathbf{0},\, \ldots,\, \mathbf{0}\big]
\in \mathbb{R}^{C \times T \times H' \times W'}.
\end{equation}
The transformer input is the channel concatenation
\begin{equation}
\mathbf{z}_{\text{in}} = [\mathbf{z}^{(\tau)};\mathbf{z}_h;\mathbf{z}_r] \in \mathbb{R}^{3C \times T \times H' \times W'}.
\end{equation}

\subsection{Camera Control Injection}

\noindent\textbf{Temporal packing.}
Per-frame Pl\"ucker-ray maps are $P_t \in \mathbb{R}^{6 \times H \times W}$.
Since the VAE compresses time by $4\times$, each latent time index aggregates four consecutive frames.
Let $\ell \in \{0,\ldots,T-1\}$ denote the latent-time index.
We pack four consecutive Pl\"ucker-ray maps into a 24-channel tensor:
\begin{equation}
P^{(\ell)}_{\text{pack}} =
\big[P_{4\ell};\,P_{4\ell+1};\,P_{4\ell+2};\,P_{4\ell+3}\big]
\in \mathbb{R}^{24 \times H \times W},
\end{equation}
with edge padding by repeating the last available frame when $4\ell{+}k$ exceeds the sequence length.

\noindent\textbf{Adapter architecture.}
A lightweight camera adapter $a_{\text{cam}}$ maps $P_{\text{pack}}$ to the patch-token grid.
Concretely, it applies PixelUnshuffle($8$) for spatial compression, a Conv2D layer aligned with the patch size/stride, and a residual block to output token-aligned embeddings.

\noindent\textbf{Injection.}
Camera embeddings are added to patch tokens:
\begin{equation}
\mathbf{h}_0 = \mathrm{Emb}_{\text{patch}}(\mathbf{z}_{\text{in}}) + a_{\text{cam}}(P_{\text{pack}}),
\end{equation}
which empirically yields more stable viewpoint conditioning than channel concatenation.

\subsection{LoRA Fine-tuning}

We apply LoRA~\cite{lora} (rank 256, $\alpha{=}256$) to the query, key, value projections and the feed-forward layers of all transformer blocks.
This enables efficient adaptation while preserving the pretrained prior of Wan2.1-1.3B-Control.

\subsection{Training Schedule and Hyperparameters}

Training uses bf16 mixed precision on 8$\times$ NVIDIA A100-80GB GPUs.

\noindent\textbf{Stage 1: camera-adapter pretraining.}
We train only the camera adapter with the transformer backbone frozen for $\sim$10K steps using learning rate $1{\times}10^{-4}$.

\noindent\textbf{Stage 2: joint fine-tuning.}
We jointly fine-tune the camera adapter and LoRA parameters for $\sim$100K iterations.
We use AdamW with learning rate $1{\times}10^{-4}$, $(b_1,b_2)=(0.9,0.999)$, weight decay 0.03, $\varepsilon=10^{-10}$, and gradient clipping 0.05.
We sample 41-frame clips at 480p (short side, aspect ratio preserved) with batch size 8 (1 per GPU).
We retain training samples where at least 30\% of frames contain non-empty hand control signals.

\subsection{Inference Settings}

\noindent\textbf{Bidirectional diffusion teacher.}
We use a flow-matching Euler scheduler with 50 sampling steps, set the text prompt to empty, and generate 81-frame sequences at 480p with original aspect ratio.

\noindent\textbf{AR generator.}
The distilled AR generator produces $B{=}3$ frames per block using $N_s{=}4$ denoising steps, and uses KV caching to keep per-block cost approximately constant as the rollout length increases.
We set the text prompt to empty and generate sequences at 384p with original aspect ratio.

\section{Auto-Annotation Pipeline}
\label{sec:supp_pipeline}

Training requires paired hand geometry and camera motion signals, which are typically unavailable for in-the-wild egocentric videos.
We therefore recover both from monocular videos and use them as pseudo-labels for conditioning.

\subsection{Hand Control Signal Extraction}

We extract hand controls through detection, temporal stabilization, mesh reconstruction, and rendering.

\noindent\textbf{Detection.}
A YOLO-based hand detector predicts per-frame bounding boxes with left/right classification.
We retain the highest-confidence detection for each class per frame and enlarge boxes by $1.2\times$ to provide context for reconstruction.

\noindent\textbf{Temporal stabilization (rule-based).}
We apply lightweight heuristics that are fast enough for large-scale processing:
(i) \emph{Overlap filtering}: when left/right detections have IoU $>0.5$, we retain only the higher-confidence one.
(ii) \emph{Short-gap interpolation}: we linearly interpolate brief missing segments when re-appearance is spatially close to the last valid position.
(iii) \emph{Edge filter}: interpolation is skipped when either endpoint box falls within 10\% of the image boundary to avoid spurious extrapolation near entry/exit events.
After interpolation, we re-run the overlap filter to remove newly introduced duplicates.

\noindent\textbf{Mesh reconstruction.}
We run HaMeR~\cite{HaMeR} per frame to estimate MANO~\cite{MANO2017} parameters and obtain per-hand vertices $V_t^h \in \mathbb{R}^{778\times 3}$.

\noindent\textbf{Projection and rendering.}
We project meshes using estimated intrinsics and render a composite control signal consisting of a filled silhouette plus a wireframe overlay.
Left/right hands are color-coded to preserve identity under overlap.
This representation specifies intended geometry while leaving visibility/occlusion to be inferred from the scene context.

\subsection{Camera Pose Estimation and Pl\"ucker-Ray Embeddings}

We estimate per-frame camera parameters using Depth Anything V3~\cite{depthanything3} in streaming mode, which processes videos in overlapping chunks and aligns consecutive chunks via SIM3 transformations computed over the shared overlap region.
During training, we use DA3-Nested-Giant-Large-1.1 with chunk size 120 and overlap 60 for maximum accuracy. At inference time, we switch to DA3-Base with chunk size 12 and overlap 3 (yielding 9 effective output frames per chunk) for speed.
We normalize trajectories relative to the first frame before computing Pl\"ucker-ray coordinates.
Depth is used only for pose recovery and is not provided to the generator.

\section{Autoregressive Distillation Details}
\label{sec:supp_ar}

We distill the bidirectional diffusion teacher into a causal AR generator following CausVid~\cite{yin2025causvid} with self-forcing~\cite{huang2025selfforcing}.

\subsection{ODE Pretraining}

We generate teacher trajectories by solving the probability-flow ODE from noise to data and train the student to match these trajectories.
This initializes the student before distribution matching.

\subsection{Asymmetric Distribution Matching}

The AR generator produces blocks autoregressively. Let $G_{\theta}^{\text{AR}}$ denote the AR generator:
\begin{equation}
\hat{I}_{t:t+B} = G_{\theta}^{\text{AR}}(\{I_1,\ldots,I_{t-1}\}, S_{t:t+B}, C_{t:t+B}),
\end{equation}
where $B{=}3$ and $S_{t:t+B}$ denotes the rendered hand-control frames for the current block.
The frozen teacher supervises via distribution matching distillation (DMD)~\cite{yin2024improved}:
\begin{equation}
\mathcal{L}_{\text{DMD}} =
\mathbb{E}_{t,\tau}\Big[\|\mathbf{v}_{\text{student}}(\hat{I}_t,\tau)-\mathbf{v}_{\text{teacher}}(I_t,\tau)\|_2^2\Big],
\end{equation}
where $\mathbf{v}$ denotes the velocity field prediction.
We freeze the camera adapter during distillation to preserve learned viewpoint conditioning.

\subsection{Self-Forcing}

Self-forcing reduces exposure bias by occasionally replacing ground-truth history with student-generated history during training:
\begin{equation}
\tilde{I}_t =
\begin{cases}
I_t^{\text{GT}} & \text{with probability } 1-p,\\
\hat{I}_t^{\text{student}} & \text{with probability } p,
\end{cases}
\end{equation}
where $p$ is annealed from 0 to 0.5.
This improves stability for long-horizon rollouts.

\subsection{Inference with KV Caching}

The AR generator produces video block-wise and caches key/value states from previous blocks as context.
This supports arbitrary-length rollouts with approximately constant throughput per block.

\section{Evaluation Protocol and Additional Results}
\label{sec:supp_eval}

This section specifies datasets, metric definitions, and extended quantitative results.

\subsection{Datasets}

We evaluate on three egocentric hand--object interaction datasets spanning complementary conditions.
Table~\ref{tab:supp_dataset_comparison} summarizes their properties.

\begin{table}[t]
\centering
\caption{Properties of datasets used for evaluation. HOT3D provides realistic head motion from augmented-reality glasses. ARCTIC emphasizes objects with state changes. HOI4D contributes category-level object diversity.}
\label{tab:supp_dataset_comparison}
\resizebox{\textwidth}{!}{%
\begin{tabular}{lcccc}
\toprule
\textbf{Dataset} & \textbf{Object Types} & \textbf{Interaction Complexity} & \textbf{Scene Diversity} & \textbf{Camera Motion} \\
\midrule
HOT3D~\cite{Banerjee_2025_CVPR} & 33 rigid household objects & Daily manipulation tasks & Varied indoor layouts & Large-amplitude head motion \\
 & (mugs, bottles, tools) & (kitchen, office, living room) & with diverse lighting & (natural glasses-wearer motion) \\
\midrule
ARCTIC~\cite{fan2023arctic} & 11 objects & Bimanual dexterous manipulation & Controlled MoCap environment & Predominantly static viewpoints \\
 & (scissors, laptops, phones) & with state changes (open/close) & with consistent background & (8 fixed + 1 egocentric camera) \\
\midrule
HOI4D~\cite{Liu_2022_CVPR} & 800+ objects (16 categories) & Functionality-oriented tasks & 610 distinct indoor rooms & Moderate egocentric motion \\
 & (rigid and articulated) & (54 distinct task types) & with cluttered backgrounds & (task-focused movement) \\
\bottomrule
\end{tabular}%
}
\end{table}

For evaluation, we segment test sequences into 81-frame clips.
Before computing metrics, we resize generated and ground-truth clips to the same spatial resolution (short side 480 by default, with aspect ratio preserved) to ensure consistent evaluation.

\subsection{Metrics}
\label{sec:supp_metrics}

We report eight metrics for realism, fidelity, temporal coherence, and 3D consistency.

\noindent\textbf{FVD.}
Fr\'echet Video Distance (FVD)~\cite{styleganv} measures distributional similarity using I3D features (lower is better).

\noindent\textbf{DINO.}
DINO similarity~\cite{dino} measures semantic alignment using ViT features (higher is better).

\noindent\textbf{PSNR, SSIM, and LPIPS.}
We report PSNR, SSIM~\cite{wang2004ssim}, and LPIPS~\cite{lpips} for frame-wise fidelity.

\noindent\textbf{Flow-ERR.}
We compute optical flow using RAFT~\cite{raft} and report an average flow error between generated and ground-truth sequences (lower is better).

\noindent\textbf{Depth-ERR and Cam-ERR.}
We estimate depth and camera trajectories using Depth Anything V3~\cite{depthanything3}.
Depth-ERR measures depth-map discrepancy, and Cam-ERR measures viewpoint consistency via per-pixel Pl\"ucker-ray embeddings (lower is better for both).

\subsection{Quantitative Comparison with Baselines}

Table~\ref{tab:supp_per_dataset} reports per-dataset results for Hand2World (bidirectional teacher) and Hand2World-AR (distilled AR generator) together with four monocular baselines. Hand2World ranks first across all metrics on all three datasets. Hand2World-AR remains close to the teacher while consistently outperforming prior methods.

\noindent\textbf{ARCTIC.}
ARCTIC emphasizes articulated objects with state changes, requiring the model to track geometry evolution over time.
Hand2World achieves an FVD of 218.76, substantially outperforming all baselines.
InterDyn achieves the best baseline FVD at 908.32 by conditioning on dense hand masks that provide per-frame spatial structure, yet it lacks explicit camera modeling, leading to viewpoint drift and elevated Depth-ERR of 27.23.
Mask2IV requires pre-specified object trajectory masks as input, constraining its generative flexibility for dynamic state changes.
CosHand generates individual images rather than temporally coherent video, which limits its ability to model continuous state transitions.
The backbone Wan2.1-1.3B-Control, without dedicated hand or camera conditioning, cannot effectively control the interaction dynamics.
By projecting the full 3D hand mesh and injecting Pl\"ucker-ray camera embeddings, Hand2World preserves articulation detail regardless of occlusion while disentangling viewpoint from hand motion, reducing Cam-ERR to 0.07 from the baseline range of 0.12--0.14.

\noindent\textbf{HOT3D.}
HOT3D features large-amplitude head motion and stresses viewpoint control.
None of the baselines include explicit camera modeling, which produces floating backgrounds under large ego-motion.
The backbone Wan2.1-1.3B-Control achieves the best baseline FVD at 349.89, while mask-based methods InterDyn and Mask2IV achieve higher FVD of 540.76 and 550.49---their 2D masks shift with head motion, introducing additional spatial inconsistency beyond the lack of camera control.
Hand2World reduces FVD to 106.20.
The benefit of explicit camera conditioning is most visible in Cam-ERR: all baselines remain between 0.33 and 0.38, reflecting uncorrected viewpoint drift, while Hand2World achieves 0.13.
Flow-ERR and Depth-ERR improve in parallel, confirming that Pl\"ucker-ray injection stabilizes both background geometry and temporal dynamics under large head motion.

\noindent\textbf{HOI4D.}
HOI4D introduces diverse object categories across 610 rooms, stressing generalization in cluttered environments.
Hand2World achieves an FVD of 251.05, outperforming all baselines.
The diversity of scenes highlights different characteristics of each conditioning strategy.
CosHand attains competitive per-frame PSNR of 18.60 through its mask-based conditioning but generates individual images without temporal modeling, yielding an FVD of 1090.31.
InterDyn and Mask2IV benefit from spatial mask conditioning and achieve moderate Depth-ERR of 14.19 and 13.96, but without camera control their perceptual quality remains limited, with LPIPS of 0.38 and 0.44 compared to 0.19 for Hand2World.
Hand2World maintains both perceptual quality and viewpoint consistency, achieving Cam-ERR of 0.04 and Depth-ERR of 7.98, benefiting from the object-agnostic nature of 3D hand mesh conditioning and explicit camera control.

\noindent\textbf{Teacher vs.\ student.}
At the 81-frame horizon, Hand2World-AR remains close to the bidirectional teacher across datasets.
For example, on HOT3D, FVD increases from 106.20 to 112.90, while Cam-ERR changes only from 0.13 to 0.14.
Across ARCTIC and HOI4D, the AR generator preserves camera consistency to within 0.01 Cam-ERR of the teacher, indicating that causal distillation largely retains both visual quality and viewpoint control.

\begin{table}[!t]
\centering
\caption{Quantitative comparison on ARCTIC, HOT3D, and HOI4D test sets. Best results are \textbf{bold}.}
\label{tab:supp_per_dataset}
\resizebox{\textwidth}{!}{%
\begin{tabular}{llccccccccc}
\toprule
\textbf{Dataset} & \textbf{Method} & \textbf{FVD}$\downarrow$ & \textbf{DINO}$\uparrow$ & \textbf{PSNR}$\uparrow$ & \textbf{SSIM}$\uparrow$ & \textbf{LPIPS}$\downarrow$ & \textbf{Flow-ERR}$\downarrow$ & \textbf{Depth-ERR}$\downarrow$ & \textbf{Cam-ERR}$\downarrow$ \\
\midrule
\multirow{6}{*}{ARCTIC}
& CosHand~\cite{coshand} & 1400.68 & 0.79 & 12.41 & 0.63 & 0.55 & 78.87 & 22.51 & 0.14 \\
& InterDyn~\cite{interdyn} & 908.32 & 0.80 & 12.43 & 0.63 & 0.53 & 72.20 & 27.23 & 0.13 \\
& Mask2IV~\cite{li2026mask2iv} & 1083.62 & 0.78 & 11.47 & 0.62 & 0.55 & 70.45 & 31.34 & 0.12 \\
& Wan2.1-1.3B-Control~\cite{wan2025} & 1038.53 & 0.75 & 12.81 & 0.62 & 0.49 & 63.71 & 22.66 & 0.13 \\
\cmidrule{2-10}
& \cellcolor{yellow!30}\textbf{Hand2World-AR (Ours)} & \cellcolor{yellow!30}232.40 & \cellcolor{yellow!30}\textbf{0.88} & \cellcolor{yellow!30}14.55 & \cellcolor{yellow!30}\textbf{0.67} & \cellcolor{yellow!30}0.41 & \cellcolor{yellow!30}46.10 & \cellcolor{yellow!30}17.02 & \cellcolor{yellow!30}\textbf{0.07} \\
& \cellcolor{yellow!30}\textbf{Hand2World (Ours)} & \cellcolor{yellow!30}\textbf{218.76} & \cellcolor{yellow!30}\textbf{0.88} & \cellcolor{yellow!30}\textbf{14.93} & \cellcolor{yellow!30}\textbf{0.67} & \cellcolor{yellow!30}\textbf{0.39} & \cellcolor{yellow!30}\textbf{44.43} & \cellcolor{yellow!30}\textbf{16.14} & \cellcolor{yellow!30}\textbf{0.07} \\
\midrule
\multirow{6}{*}{HOT3D}
& CosHand~\cite{coshand} & 786.07 & 0.84 & 15.97 & 0.50 & 0.47 & 82.72 & 35.74 & 0.34 \\
& InterDyn~\cite{interdyn} & 540.76 & 0.81 & 13.47 & 0.42 & 0.58 & 68.46 & 43.22 & 0.38 \\
& Mask2IV~\cite{li2026mask2iv} & 550.49 & 0.77 & 13.11 & 0.35 & 0.61 & 71.99 & 32.76 & 0.37 \\
& Wan2.1-1.3B-Control~\cite{wan2025} & 349.89 & 0.78 & 15.87 & 0.51 & 0.47 & 68.81 & 24.73 & 0.33 \\
\cmidrule{2-10}
& \cellcolor{yellow!30}\textbf{Hand2World-AR (Ours)} & \cellcolor{yellow!30}112.90 & \cellcolor{yellow!30}0.89 & \cellcolor{yellow!30}18.25 & \cellcolor{yellow!30}0.57 & \cellcolor{yellow!30}0.34 & \cellcolor{yellow!30}41.85 & \cellcolor{yellow!30}15.59 & \cellcolor{yellow!30}0.14 \\
& \cellcolor{yellow!30}\textbf{Hand2World (Ours)} & \cellcolor{yellow!30}\textbf{106.20} & \cellcolor{yellow!30}\textbf{0.91} & \cellcolor{yellow!30}\textbf{18.79} & \cellcolor{yellow!30}\textbf{0.59} & \cellcolor{yellow!30}\textbf{0.32} & \cellcolor{yellow!30}\textbf{40.06} & \cellcolor{yellow!30}\textbf{15.37} & \cellcolor{yellow!30}\textbf{0.13} \\
\midrule
\multirow{6}{*}{HOI4D}
& CosHand~\cite{coshand} & 1090.31 & 0.87 & 18.60 & 0.73 & 0.35 & 81.88 & 21.97 & 0.11 \\
& InterDyn~\cite{interdyn} & 668.35 & 0.87 & 16.91 & 0.70 & 0.38 & 61.24 & 14.19 & 0.08 \\
& Mask2IV~\cite{li2026mask2iv} & 710.92 & 0.86 & 15.43 & 0.69 & 0.44 & 67.83 & 13.96 & 0.09 \\
& Wan2.1-1.3B-Control~\cite{wan2025} & 603.25 & 0.80 & 17.07 & 0.70 & 0.37 & 60.26 & 17.50 & 0.15 \\
\cmidrule{2-10}
& \cellcolor{yellow!30}\textbf{Hand2World-AR (Ours)} & \cellcolor{yellow!30}263.40 & \cellcolor{yellow!30}0.91 & \cellcolor{yellow!30}20.95 & \cellcolor{yellow!30}0.75 & \cellcolor{yellow!30}0.21 & \cellcolor{yellow!30}38.90 & \cellcolor{yellow!30}8.55 & \cellcolor{yellow!30}0.05 \\
& \cellcolor{yellow!30}\textbf{Hand2World (Ours)} & \cellcolor{yellow!30}\textbf{251.05} & \cellcolor{yellow!30}\textbf{0.93} & \cellcolor{yellow!30}\textbf{21.57} & \cellcolor{yellow!30}\textbf{0.76} & \cellcolor{yellow!30}\textbf{0.19} & \cellcolor{yellow!30}\textbf{37.25} & \cellcolor{yellow!30}\textbf{7.98} & \cellcolor{yellow!30}\textbf{0.04} \\
\bottomrule
\end{tabular}%
}
\end{table}

\subsection{Quantitative Ablation Studies}

Table~\ref{tab:supp_ablation_full} reports ablation on ARCTIC, HOT3D, and HOI4D. We evaluate explicit camera conditioning via the camera adapter, the wireframe overlay used in the hand-control rendering, and temporal stabilization in our monocular auto-annotation pipeline.

\noindent\textbf{Explicit camera conditioning is the primary contributor.}
Removing the camera adapter yields the largest degradation across all datasets. FVD increases from 218.76 to 815.14 on ARCTIC, from 106.20 to 438.82 on HOT3D, and from 251.05 to 632.58 on HOI4D, corresponding to 3.7$\times$, 4.1$\times$, and 2.5$\times$ higher FVD.
This change also substantially harms 3D consistency. Cam-ERR increases from 0.07 to 0.13 on ARCTIC, from 0.13 to 0.29 on HOT3D, and from 0.04 to 0.07 on HOI4D.
Depth-ERR increases from 15.37 to 23.51 on HOT3D and from 7.98 to 10.42 on HOI4D. Temporal and perceptual metrics deteriorate in parallel, with HOT3D Flow-ERR increasing from 40.06 to 64.42 and LPIPS increasing from 0.32 to 0.44.
Overall, the results show that an explicit camera pathway is critical for disentangling viewpoint change from interaction dynamics, which stabilizes background geometry under head motion.

\noindent\textbf{Wireframe rendering provides consistent but modest gains.}
Removing the wireframe overlay leads to a smaller yet consistent drop. FVD increases from 218.76 to 223.29 on ARCTIC, from 106.20 to 108.05 on HOT3D, and from 251.05 to 255.15 on HOI4D.
DINO similarity and PSNR also decrease slightly, as shown by HOT3D dropping from 0.91 to 0.90 and from 18.79 to 18.58.
These trends suggest that the wireframe complements the silhouette by exposing articulation structure, improving controllability for finger configurations beyond the hand outline.
The benefit is most evident for self-occluded poses such as palm-facing or fist configurations, where the silhouette boundary alone is ambiguous across different finger states and the wireframe edges provide structural cues to resolve the intended articulation.
We note that on HOT3D, Depth-ERR is marginally lower without the wireframe (15.35 vs.\ 15.37); this difference is within noise and does not affect the overall trend.

\noindent\textbf{Temporal stabilization is mainly driven by short-gap interpolation.}
Among the stabilization components, short-gap interpolation contributes the most. Disabling interpolation increases FVD from 218.76 to 231.09 on ARCTIC, from 106.20 to 111.19 on HOT3D, and from 251.05 to 265.47 on HOI4D.
In contrast, removing edge filtering or overlap filtering has a minor effect, with FVD changing by at most 0.5\% across all datasets. This matches their role in suppressing occasional boundary jitter and duplicate detections during large-scale auto-annotation.

\begin{table}[!t]
\centering
\caption{Complete ablation study across ARCTIC, HOT3D, and HOI4D. We ablate the camera adapter (Pl\"ucker-ray injection), wireframe overlay, temporal interpolation, boundary (edge) filtering, and overlap filtering. Best results are \textbf{bold}.}
\label{tab:supp_ablation_full}
\resizebox{\textwidth}{!}{%
\begin{tabular}{llcccccccc}
\toprule
\textbf{Dataset} & \textbf{Configuration} & \textbf{FVD}$\downarrow$ & \textbf{DINO}$\uparrow$ & \textbf{PSNR}$\uparrow$ & \textbf{SSIM}$\uparrow$ & \textbf{LPIPS}$\downarrow$ & \textbf{Flow-ERR}$\downarrow$ & \textbf{Depth-ERR}$\downarrow$ & \textbf{Cam-ERR}$\downarrow$ \\
\midrule
\multirow{6}{*}{ARCTIC}
& w/o Camera Adapter & 815.14 & 0.84 & 12.66 & 0.62 & 0.47 & 64.64 & 21.85 & 0.13 \\
& w/o Wireframe & 223.29 & 0.87 & 14.78 & \textbf{0.67} & \textbf{0.39} & 44.86 & \textbf{16.14} & 0.08 \\
& w/o Interpolation & 231.09 & 0.87 & 14.75 & \textbf{0.67} & \textbf{0.39} & 45.02 & 16.48 & 0.08 \\
& w/o Edge Filter & 219.15 & 0.87 & 14.76 & \textbf{0.67} & \textbf{0.39} & 44.88 & 16.25 & 0.08 \\
& w/o Overlap Filter & 219.75 & 0.87 & 14.78 & \textbf{0.67} & \textbf{0.39} & 44.94 & 16.35 & 0.08 \\
& \cellcolor{yellow!30}\textbf{Full Model} & \cellcolor{yellow!30}\textbf{218.76} & \cellcolor{yellow!30}\textbf{0.88} & \cellcolor{yellow!30}\textbf{14.93} & \cellcolor{yellow!30}\textbf{0.67} & \cellcolor{yellow!30}\textbf{0.39} & \cellcolor{yellow!30}\textbf{44.43} & \cellcolor{yellow!30}\textbf{16.14} & \cellcolor{yellow!30}\textbf{0.07} \\
\midrule
\multirow{6}{*}{HOT3D}
& w/o Camera Adapter & 438.82 & 0.86 & 16.36 & 0.51 & 0.44 & 64.42 & 23.51 & 0.29 \\
& w/o Wireframe & 108.05 & 0.90 & 18.58 & \textbf{0.59} & \textbf{0.32} & 40.13 & \textbf{15.35} & \textbf{0.13} \\
& w/o Interpolation & 111.19 & 0.90 & 18.52 & \textbf{0.59} & \textbf{0.32} & 40.28 & 15.65 & \textbf{0.13} \\
& w/o Edge Filter & 106.29 & 0.89 & 18.56 & \textbf{0.59} & 0.33 & 40.22 & 15.45 & \textbf{0.13} \\
& w/o Overlap Filter & 106.47 & 0.90 & 18.60 & \textbf{0.59} & \textbf{0.32} & 40.39 & 15.57 & \textbf{0.13} \\
& \cellcolor{yellow!30}\textbf{Full Model} & \cellcolor{yellow!30}\textbf{106.20} & \cellcolor{yellow!30}\textbf{0.91} & \cellcolor{yellow!30}\textbf{18.79} & \cellcolor{yellow!30}\textbf{0.59} & \cellcolor{yellow!30}\textbf{0.32} & \cellcolor{yellow!30}\textbf{40.06} & \cellcolor{yellow!30}15.37 & \cellcolor{yellow!30}\textbf{0.13} \\
\midrule
\multirow{6}{*}{HOI4D}
& w/o Camera Adapter & 632.58 & 0.92 & 19.83 & 0.74 & 0.26 & 46.17 & 10.42 & 0.07 \\
& w/o Wireframe & 255.15 & \textbf{0.93} & 21.48 & \textbf{0.76} & 0.20 & 37.69 & \textbf{7.98} & \textbf{0.04} \\
& w/o Interpolation & 265.47 & 0.92 & 21.34 & 0.75 & 0.20 & 37.58 & 8.10 & \textbf{0.04} \\
& w/o Edge Filter & 251.15 & 0.92 & 21.27 & \textbf{0.76} & 0.20 & 37.56 & 8.08 & \textbf{0.04} \\
& w/o Overlap Filter & 252.06 & 0.92 & 21.40 & \textbf{0.76} & 0.20 & 37.65 & 8.09 & \textbf{0.04} \\
& \cellcolor{yellow!30}\textbf{Full Model} & \cellcolor{yellow!30}\textbf{251.05} & \cellcolor{yellow!30}\textbf{0.93} & \cellcolor{yellow!30}\textbf{21.57} & \cellcolor{yellow!30}\textbf{0.76} & \cellcolor{yellow!30}\textbf{0.19} & \cellcolor{yellow!30}\textbf{37.25} & \cellcolor{yellow!30}\textbf{7.98} & \cellcolor{yellow!30}\textbf{0.04} \\
\bottomrule
\end{tabular}%
}
\end{table}

Table~\ref{tab:supp_ar_length} evaluates the AR generator beyond the 81-frame training horizon.
Hand2World-AR matches the teacher closely at 81 frames and degrades gradually as the rollout length increases.

\begin{table}[!t]
\centering
\caption{Autoregressive generation across ARCTIC, HOT3D, and HOI4D. Hand2World-AR matches the teacher closely at 81 frames and degrades gradually as the rollout length increases.}
\label{tab:supp_ar_length}
\resizebox{\textwidth}{!}{%
\begin{tabular}{lllcccccccc}
\toprule
\textbf{Dataset} & \textbf{Attention} & \textbf{Len} & \textbf{FVD}$\downarrow$ & \textbf{DINO}$\uparrow$ & \textbf{PSNR}$\uparrow$ & \textbf{SSIM}$\uparrow$ & \textbf{LPIPS}$\downarrow$ & \textbf{Flow-ERR}$\downarrow$ & \textbf{Depth-ERR}$\downarrow$ & \textbf{Cam-ERR}$\downarrow$ \\
\midrule
\multirow{4}{*}{ARCTIC}
& Bidirectional & 81 & \textbf{218.76} & \textbf{0.88} & \textbf{14.93} & \textbf{0.67} & \textbf{0.39} & \textbf{44.43} & \textbf{16.14} & \textbf{0.07} \\
& Autoregressive & 81 & 232.40 & \textbf{0.88} & 14.55 & \textbf{0.67} & 0.41 & 46.10 & 17.02 & \textbf{0.07} \\
& Autoregressive & 162 & 264.20 & \textbf{0.88} & 13.53 & 0.66 & 0.48 & 49.01 & 18.72 & 0.12 \\
& Autoregressive & 324 & 331.71 & 0.86 & 12.45 & 0.63 & 0.52 & 54.71 & 20.99 & 0.17 \\
\midrule
\multirow{4}{*}{HOT3D}
& Bidirectional & 81 & \textbf{106.20} & \textbf{0.91} & \textbf{18.79} & \textbf{0.59} & \textbf{0.32} & \textbf{40.06} & \textbf{15.37} & \textbf{0.13} \\
& Autoregressive & 81 & 112.90 & 0.89 & 18.25 & 0.57 & 0.34 & 41.85 & 15.59 & 0.14 \\
& Autoregressive & 162 & 292.73 & 0.86 & 16.98 & 0.55 & 0.42 & 47.26 & 16.10 & 0.16 \\
& Autoregressive & 324 & 418.64 & 0.80 & 15.40 & 0.50 & 0.50 & 54.46 & 18.82 & 0.27 \\
\midrule
\multirow{4}{*}{HOI4D}
& Bidirectional & 81 & \textbf{251.05} & \textbf{0.93} & \textbf{21.57} & \textbf{0.76} & \textbf{0.19} & \textbf{37.25} & \textbf{7.98} & \textbf{0.04} \\
& Autoregressive & 81 & 263.40 & 0.91 & 20.95 & 0.75 & 0.21 & 38.90 & 8.55 & 0.05 \\
& Autoregressive & 162 & 559.80 & 0.89 & 19.47 & 0.74 & 0.29 & 46.81 & 9.52 & 0.06 \\
& Autoregressive & 324 & 805.34 & 0.82 & 17.46 & 0.73 & 0.37 & 54.23 & 12.57 & 0.10 \\
\bottomrule
\end{tabular}%
}
\end{table}

\section{Runtime Analysis}
\label{sec:supp_runtime}

We benchmark inference speed on 324-frame sequences at $544{\times}384$ resolution using a single A100-80GB GPU.
Hand reconstruction (HaMeR) and camera estimation (Depth Anything V3) run in parallel.
The AR generator uses KV caching to keep throughput approximately constant as the rollout length increases.
Table~\ref{tab:supp_runtime} reports the runtime breakdown.

\begin{table}[h!]
\centering
\caption{\textbf{Runtime breakdown of the full Hand2World pipeline.} Throughput is measured in frames per second (FPS) on 324-frame sequences at $544{\times}384$ resolution using a single A100-80GB GPU. Hand reconstruction and camera estimation run in parallel. The AR generator is the throughput bottleneck.}
\label{tab:supp_runtime}
\begin{tabular}{ccc}
\toprule
\makecell{\textbf{Hand Reconstruction} \textbf{+ Camera Estimation}} & \textbf{AR Generator} & \textbf{Overall} \\
\midrule
25.0 & 13.7 & 8.9 \\
\bottomrule
\end{tabular}
\end{table}

\section{Qualitative Results}
\label{sec:supp_qual}

Each figure shows the reference scene image (first column) and representative frames (subsequent columns).
Rows correspond to methods. GT is the ground-truth video.
We highlight three recurring failure modes aligned with the main paper motivation:
(i) \emph{state/interaction inconsistency} (incorrect object state, unstable contacts, or appearance collapse),
(ii) \emph{viewpoint inconsistency under head motion} (background drift, incorrect rotation, or photometric instability), and
(iii) \emph{hand pose misalignment} (generated hands that deviate from the input gesture trajectory or fall out of temporal synchronization with the intended motion).

\subsection{Box opening and closing}
\label{sec:supp_qual_box}

Figures~\ref{fig:supp_recon_arctic_box_open} and~\ref{fig:supp_recon_arctic_box_close} show bimanual box manipulation, where the model must track an articulated state change---lid angle, interior visibility, and hinge geometry---over the full sequence.
This is challenging because the box interior is entirely absent from the reference image. Generating it requires a strong generative prior coupled with temporally consistent state tracking.

In Figure~\ref{fig:supp_recon_arctic_box_open}, PlayerOne produces a temporally misaligned opening progression with an increasingly shifted viewpoint from Frame~40 onward, likely due to the allocentric-to-egocentric viewpoint mismatch in its training data.
CosHand and Mask2IV introduce structural artifacts on the lid at later frames, suggesting that mask-based conditioning conflates hand motion with object state and struggles to sustain a coherent opening trajectory.
In Figure~\ref{fig:supp_recon_arctic_box_close}, PlayerOne exhibits poorly aligned hand poses and progressive viewpoint divergence at Frames~60--80, while Wan2.1-1.3B-Control degrades into severe background artifacts---without dedicated hand conditioning, the model cannot attribute scene changes to hand interaction versus viewpoint shift.
Hand2World follows the GT state transition more closely, preserving the lid angle and interior appearance throughout and maintaining stable background geometry.

\subsection{Laptop opening and closing}
\label{sec:supp_qual_laptop}

Figures~\ref{fig:supp_recon_arctic_laptop_open} and~\ref{fig:supp_recon_arctic_laptop_close} show laptop manipulation, where opening reveals entirely new visual content (keyboard and screen) that was not present in the reference image.
The thin profile of the laptop demands precise articulation control: small angular errors in the lid propagate into large visual discrepancies.

In Figure~\ref{fig:supp_recon_arctic_laptop_open}, CosHand, InterDyn, and Mask2IV fail to produce a recognizable open state, keeping the laptop surface largely flat throughout the sequence.
This suggests that 2D mask conditioning, which encodes only the visible hand extent, provides insufficient articulation cues to drive a progressive opening trajectory.
PlayerOne begins opening but shows misaligned hand positions from Frame~20 and progressive viewpoint drift.
In Figure~\ref{fig:supp_recon_arctic_laptop_close}, CosHand generates severe visual artifacts by Frame~80, and Wan2.1-1.3B-Control completely degrades by the end of the sequence.
Hand2World produces a coherent opening trajectory with visible keyboard structure at Frames~60--80, and maintains a clean closed state at the end, consistent with GT.

\subsection{Object appearance fidelity under interaction}
\label{sec:supp_qual_texture}

Figure~\ref{fig:supp_recon_hoi4d_pattern_box} shows a patterned storage box being closed.
The repeating high-frequency surface pattern is sensitive to any spatial distortion or temporal inconsistency, even subtle artifacts  might go unnoticed on a uniform surface become immediately visible on textured regions.
This case therefore serves as a stress test for both conditioning fidelity and interaction understanding.

CosHand and InterDyn gradually lose the surface pattern as the interaction progresses, with the texture washing out or shifting across frames.
This degradation is most pronounced near the hand--object boundary, where mask-based conditioning introduces spatial artifacts that propagate temporally and corrupt adjacent texture regions.
Moreover, neither method completes the closing action: the box lid remains in a similar position throughout the sequence.
Wan2.1-1.3B-Control preserves the pattern initially but produces structural distortion of the box geometry at Frames~60--80, also failing to close the box.
Hand2World is the only method that correctly closes the box while preserving both the surface pattern and box geometry, as its 3D mesh conditioning provides hand articulation cues without spatially corrupting the underlying texture.

Figure~\ref{fig:supp_recon_hot3d_cup} shows cup grasping, where the hand must grip the handle precisely and the cup moves between near and far positions relative to the camera.
This demands robust and precise hand control: the fingers must wrap around the narrow handle throughout the sequence, and the cup's identity---shape, color, and material---must remain consistent as its apparent scale changes with depth.
When the cup moves close to the camera, it occludes a large portion of the background, which the model must generate correctly upon the cup's retreat.

CosHand, InterDyn, and Wan2.1-1.3B-Control produce blurred or artifact-laden cup appearances, and the generated cup trajectories fail to match the intended near--far motion.
Mask2IV shows severe visual degradation as the cup approaches the camera.
Hand2World maintains a consistent cup identity across depth changes, renders the handle grasp faithfully, and recovers a plausible background behind the cup as it moves away.

\subsection{Large viewpoint changes}
\label{sec:supp_qual_viewpoint}

Figure~\ref{fig:supp_recon_hot3d_down} presents the same clip used in the camera ablation (Figure~\ref{fig:ablation_camera} of the main paper), now comparing all baselines.
The camera pans rightward from the initial desk view to reveal the whiteboard, then reverses leftward back toward the desk.
This non-monotonic trajectory is a stress test for viewpoint control: the model must generate new background content during the rightward sweep and maintain consistency when returning to previously seen regions.

CosHand and InterDyn remain near the initial desk view, confirming that without explicit camera conditioning these methods tend to keep the background static rather than following the intended trajectory.
Mask2IV attempts some camera movement but produces severe motion blur and visual artifacts from Frame~40 onward, with black regions appearing by Frame~80.
Wan2.1-1.3B-Control shows extreme degradation: heavy motion blur begins at Frame~40, and by Frame~60 the scene is almost entirely washed out with no recognizable structure.
Hand2World follows the rightward-then-leftward pan faithfully, revealing the whiteboard by Frames~40--60 and returning toward the desk, closely matching the GT.

\SuppFigure{figures/supp_box_open}
{Box opening. The model must generate unseen interior content as the lid opens and maintain consistent hinge geometry across frames (Sec.~\ref{sec:supp_qual_box}).}
{fig:supp_recon_arctic_box_open}

\SuppFigure{figures/supp_box_close}
{Box closing. The reverse trajectory requires tracking the lid angle from open to closed while preserving the interior appearance established during opening (Sec.~\ref{sec:supp_qual_box}).}
{fig:supp_recon_arctic_box_close}

\SuppFigure{figures/supp_laptop_open}
{Laptop opening. The thin lid profile amplifies angular errors, and the keyboard and screen must be generated as entirely new content (Sec.~\ref{sec:supp_qual_laptop}).}
{fig:supp_recon_arctic_laptop_open}

\SuppFigure{figures/supp_laptop_close}
{Laptop closing. Baselines that failed to open the laptop in Figure~\ref{fig:supp_recon_arctic_laptop_open} also fail to produce a coherent closing trajectory (Sec.~\ref{sec:supp_qual_laptop}).}
{fig:supp_recon_arctic_laptop_close}

\SuppFigure{figures/supp_pattern_box}
{Closing a patterned storage box. The high-frequency surface texture exposes any spatial distortion or temporal inconsistency, and only Hand2World completes the closing action (Sec.~\ref{sec:supp_qual_texture}).}
{fig:supp_recon_hoi4d_pattern_box}

\SuppFigure{figures/supp_towel}
{Cup grasping with near--far depth variation. Precise handle grip control and consistent cup identity across scale changes are required, along with plausible background generation behind the occluding cup (Sec.~\ref{sec:supp_qual_texture}).}
{fig:supp_recon_hot3d_cup}

\SuppFigure{figures/supp_ablation_camera}
{Large rightward-then-leftward head pan. The camera sweeps from a desk view to the whiteboard and reverses back, testing both novel-view generation and return consistency. This is the same clip as the camera ablation in Figure~\ref{fig:ablation_camera} of the main paper (Sec.~\ref{sec:supp_qual_viewpoint}).}
{fig:supp_recon_hot3d_down}

\section{Additional Qualitative Results on Challenging Cases}
\label{sec:supp_challenge_case}

Sec.~\ref{sec:supp_eval} provides complete ablation studies (Table~\ref{tab:supp_ablation_full}) and autoregressive length scaling results (Table~\ref{tab:supp_ar_length}).
Here we further visualize challenging long-horizon autoregressive rollouts in Figure~\ref{fig:supp_long}.

\subsection{Challenging Cases}

Figure~\ref{fig:supp_long} reports long-horizon rollouts on challenging scenarios.
Rows 1--2 emphasize \emph{unseen state/content generation}: the reference image contains only the closed-book state, while the rollout requires opening and maintaining page content over time.
Rows 3--4 emphasize \emph{extreme viewpoint changes and scale variation}: the camera undergoes substantial rotation and the manipulated object alternates between close-up and far views.
Across these cases, Hand2World-AR maintains coherent interaction dynamics and a viewpoint-consistent background over extended horizons.

\SuppFigure{figures/supp_challenge_case}
{Four challenging Hand2World-AR rollouts at approximately $4{\times}$ the training clip length (160 frames). Rows~1--2: book manipulation, where the model must generate unseen page content as the book is opened and flipped. Rows~3--4: cup and container interaction with extreme viewpoint rotation and large scale variation as objects move close to and away from the camera.}
{fig:supp_long}

\section{Free-Space Gesture Generation}
\label{sec:supp_freespace}

We evaluate on free-space mid-air gestures where the user's hands do not touch any object.
As discussed in Sec.~\ref{sec:hand} of the main paper, this setting introduces a train--test distribution shift for mask-based methods because free-space gestures produce complete hand masks that differ from the partial masks seen during training.
We present two examples that combine this distribution shift with additional challenges: viewpoint change (Figure~\ref{fig:supp_freespace_1}) and motion blur (Figure~\ref{fig:supp_freespace_2}).

\subsection{Free-space gestures under viewpoint change}
\label{sec:supp_freespace_viewpoint}

In Figure~\ref{fig:supp_freespace_1}, the input gesture stream (top row) drives a box-opening motion while the camera pans rightward and then reverses leftward.
This non-monotonic viewpoint trajectory tests two capabilities simultaneously: whether the generated hands remain photorealistic under the mask distribution shift, and whether the synthesized box and hand positions track the viewpoint reversal without spatial drift.

CosHand generates a box that partially opens but produces increasingly unnatural hand poses from Frame~40 onward, with the fingers losing contact with the box surface and the opening progression becoming temporally inconsistent.
InterDyn and Mask2IV generate a box but fail to produce a clear opening trajectory---the box orientation shifts across frames without a coherent lid-opening progression, and the hands drift spatially relative to the box as the viewpoint reverses.
Wan2.1-1.3B-Control produces persistent orange-tinted hand color throughout the sequence, a direct symptom of the mask distribution shift: receiving a complete hand mask at inference, the model generates unnatural skin tones it never encountered during training with occluded hands. The box remains largely closed.
Hand2World is the only method that produces a coherent opening trajectory with a visible interior at Frames~60--80, maintains natural hand appearance with consistent skin tone, and keeps the hands spatially aligned with the box throughout the viewpoint reversal.

\subsection{Free-space gestures under motion blur}
\label{sec:supp_freespace_blur}

In Figure~\ref{fig:supp_freespace_2}, the driving stream contains severe motion blur around Frames~40--80 caused by rapid head rotation, compounding the free-space distribution shift with degraded input quality.
This is a particularly challenging setting because the model must extract reliable hand articulation cues from blurred input frames while simultaneously handling a viewpoint change.

CosHand generates a box at Frame~20 but loses spatial consistency as the blur intensifies---by Frame~60 the box orientation has shifted substantially, and while the lid appears partially raised, the opening progression is inconsistent across frames.
InterDyn and Mask2IV keep the box largely closed throughout the sequence, with the hands drifting spatially and decoupling from the intended opening motion.
Without explicit camera conditioning, these methods cannot separate hand displacement from the viewpoint-induced parallax caused by the rapid head rotation.
Wan2.1-1.3B-Control keeps the box closed and exhibits persistent hand color distortion, with red-tinted skin visible throughout the sequence.
Hand2World produces a clear opening trajectory---the lid progressively rises and a plausible interior becomes visible by Frame~60---while maintaining natural hand appearance and stable background despite the blurred driving input.

\SuppFigure[0.84]{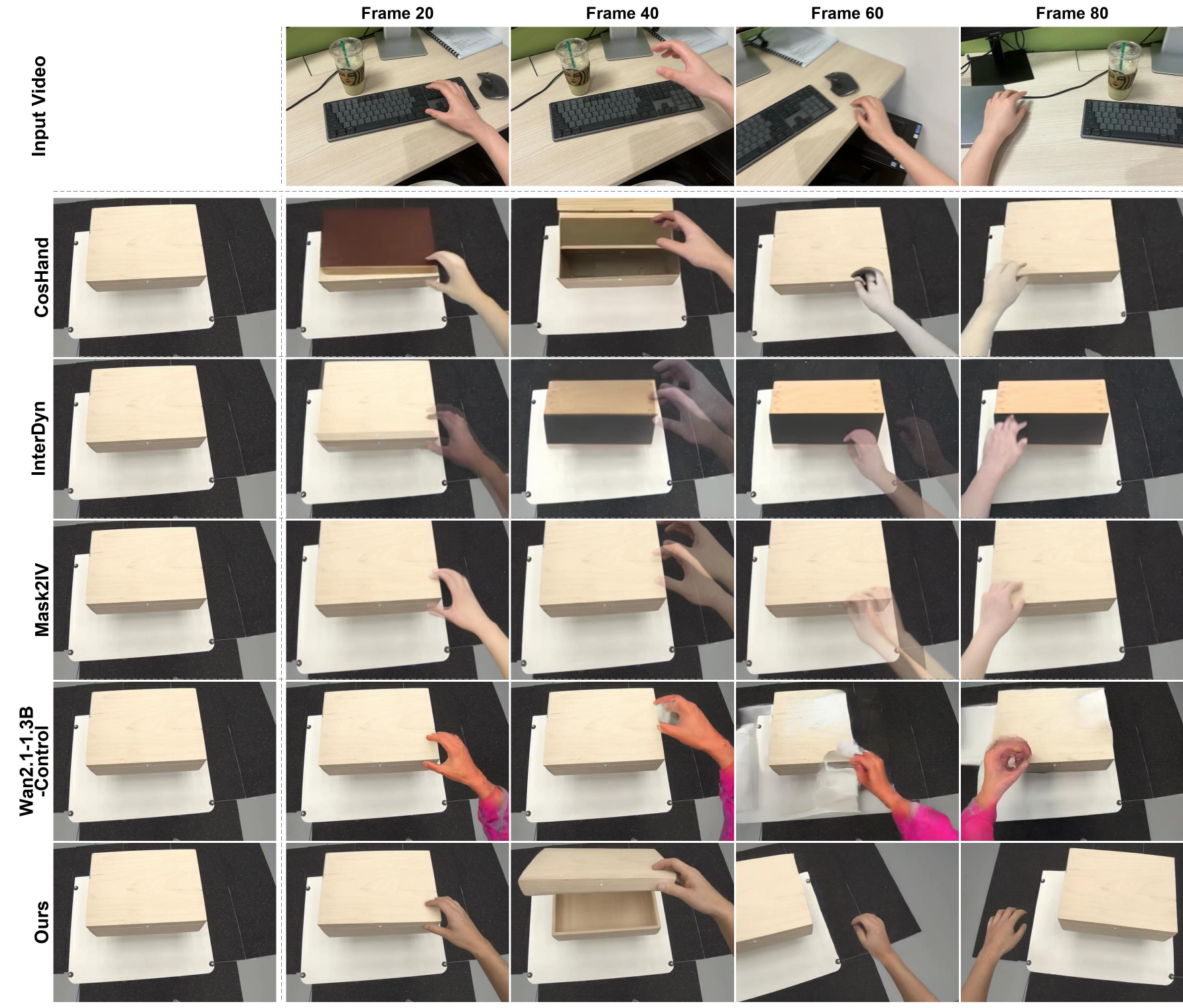}
{Free-space box opening with non-monotonic camera pan. The camera pans rightward then reverses leftward, requiring consistent hand placement and box state under a changing viewpoint trajectory (Sec.~\ref{sec:supp_freespace_viewpoint}).}
{fig:supp_freespace_1}

\SuppFigure[0.84]{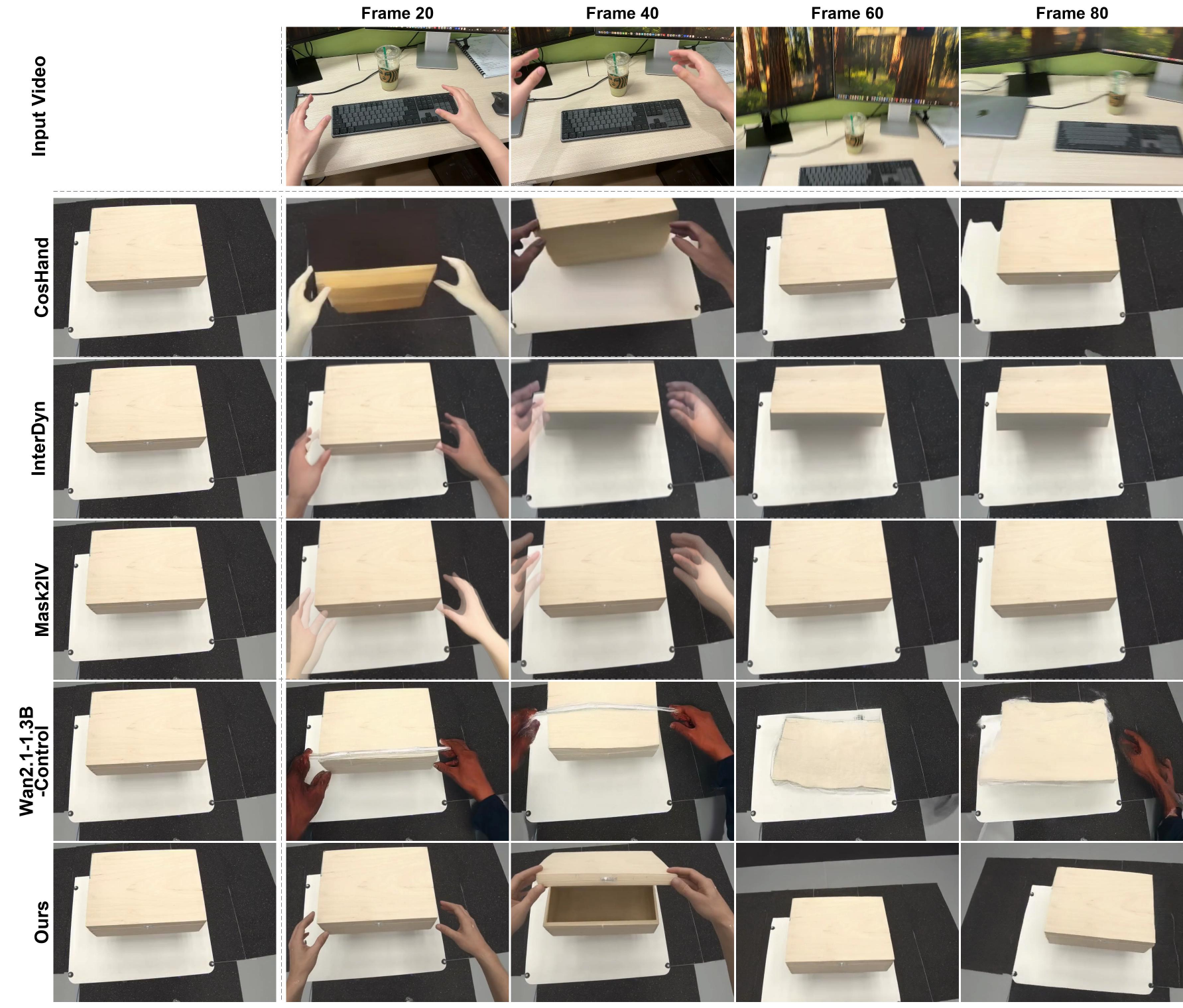}
{Free-space box opening under severe motion blur. The driving stream contains heavy blur around Frames~40--60 from rapid head rotation, compounding the distribution shift with degraded input quality (Sec.~\ref{sec:supp_freespace_blur}).}
{fig:supp_freespace_2}